\documentclass[runningheads]{llncs}

\usepackage{eccv}

\usepackage{eccvabbrv}

\usepackage{graphicx}
\usepackage{booktabs}
\usepackage{bm}
\usepackage{multirow}
\usepackage{adjustbox}

\usepackage[accsupp]{axessibility}  % Improves PDF readability for those with disabilities.

\usepackage[pagebackref,breaklinks,colorlinks,citecolor=eccvblue]{hyperref}

\usepackage{orcidlink}

\newcommand{\tikzcircle}[2][red,fill=red]{\tikz[baseline=-0.7ex]\draw[#1,radius=#2] (0,0) circle ;}

\definecolor{gold}{RGB}{221, 196, 65}
\definecolor{silver}{RGB}{215, 215, 215}
\definecolor{bronze}{RGB}{126, 66, 5}

\graphicspath{{images/}}

\definecolor{yellow}{rgb}{1, 1, 0.7}
\definecolor{orange}{rgb}{1, 0.85, 0.7}
\definecolor{pink}{rgb}{1, 0.7, 0.7}

\begin{document}

\title{Portrait4D-v2: Pseudo Multi-View Data Creates Better 4D Head Synthesizer} 

% \titlerunning{Portrait4D-v2}

\author{Yu Deng\orcidlink{0000-0001-7241-8519} \and
Duomin Wang\orcidlink{0009-0004-9507-6741} \and
Baoyuan Wang\orcidlink{0000-0002-8268-7517}}

\authorrunning{Y. Deng et al.}

\institute{Xiaobing.AI\\
\url{https://yudeng.github.io/Portrait4D-v2/}}

\maketitle

\begin{abstract}
  In this paper, we propose a novel learning approach for feed-forward one-shot 4D head avatar synthesis. Different from existing methods that often learn from reconstructing monocular videos guided by 3DMM, we employ pseudo multi-view videos to learn a 4D head synthesizer in a data-driven manner, avoiding reliance on inaccurate 3DMM reconstruction that could be detrimental to the synthesis performance. The key idea is to first learn a 3D head synthesizer using synthetic multi-view images to convert monocular real videos into multi-view ones, and then utilize the pseudo multi-view videos to learn a 4D head synthesizer via cross-view self-reenactment. By leveraging a simple vision transformer backbone with motion-aware cross-attentions, our method exhibits superior performance compared to previous methods in terms of reconstruction fidelity, geometry consistency, and motion control accuracy. We hope our method offers novel insights into integrating 3D priors with 2D supervisions for improved 4D head avatar creation.
\end{abstract}

\section{Introduction}
\label{sec:intro}
One-shot head avatar synthesis has garnered significant attention in recent years among computer vision and graphics community. Its typical problem setting is to generate photorealistic portrait videos given a source image for appearance and a series of driving motions (from videos) for animation. Over time, the technical solution has witnessed a paradigm shift from leveraging 2D generative models~\cite{siarohin2019first,burkov2020neural,wang2021one,yin2022styleheat,drobyshev2022megaportraits} to animatable 3D (\ie, 4D) synthesizers~\cite{khakhulin2022realistic,li2023one,yu2023nofa, tran2023voodoo}, where the latter maintain better geometry consistency under significant head motions and support free-view rendering favored in AR/VR scenarios. 

Different from 2D-based methods that directly generate 2D frames via neural networks, 3D approaches synthesize head avatars in 3D space and leverage camera projection from 3D to 2D to produce the final result. While the 2D methods are often learned in a self-supervised manner using monocular videos~\cite{burkov2020neural,wang2021one}, the 3D counterparts are difficult to follow a similar procedure due to the scarcity of 3D data. Learning the 3D approaches via weak supervisions from the monocular videos often turns to be highly ill-posed without resorting to proper 3D priors. Although plenty of existing methods~\cite{khakhulin2022realistic,ma2023otavatar,li2023one,yu2023nofa,li2023generalizable} utilize head models such as 3D Morphable Models (3DMMs)~\cite{paysan20093d,li2017learning,blanz1999morphable} to help alleviate this problem, the inaccuracy of monocular 3DMM reconstructors~\cite{deng2019accurate,feng2021learning} and the limited expressiveness of a 3DMM itself often restrict them from synthesizing lifelike head avatars.

Ideally, if we have access to large-scale multi-view videos, it is possible to learn an image-to-4D synthesizer in a data-driven manner following recent successes in the 3D reconstruction field~\cite{trevithick2023real,hong2023lrm,li2023instant3d,zou2023triplane}. However, collecting such large amount of data with enough diversity for learning a generalizable head synthesizer can be impractical. A recent method Portrait4D~\cite{deng2023portrait} learns a 4D GAN in advance to synthesize large-scale multi-view data for training, yet the learned 4D GAN still requires 3DMM for expression control which is less vivid than real data.

\begin{figure*}[t]
	\small
	\centering
	\includegraphics[width=0.97\textwidth]{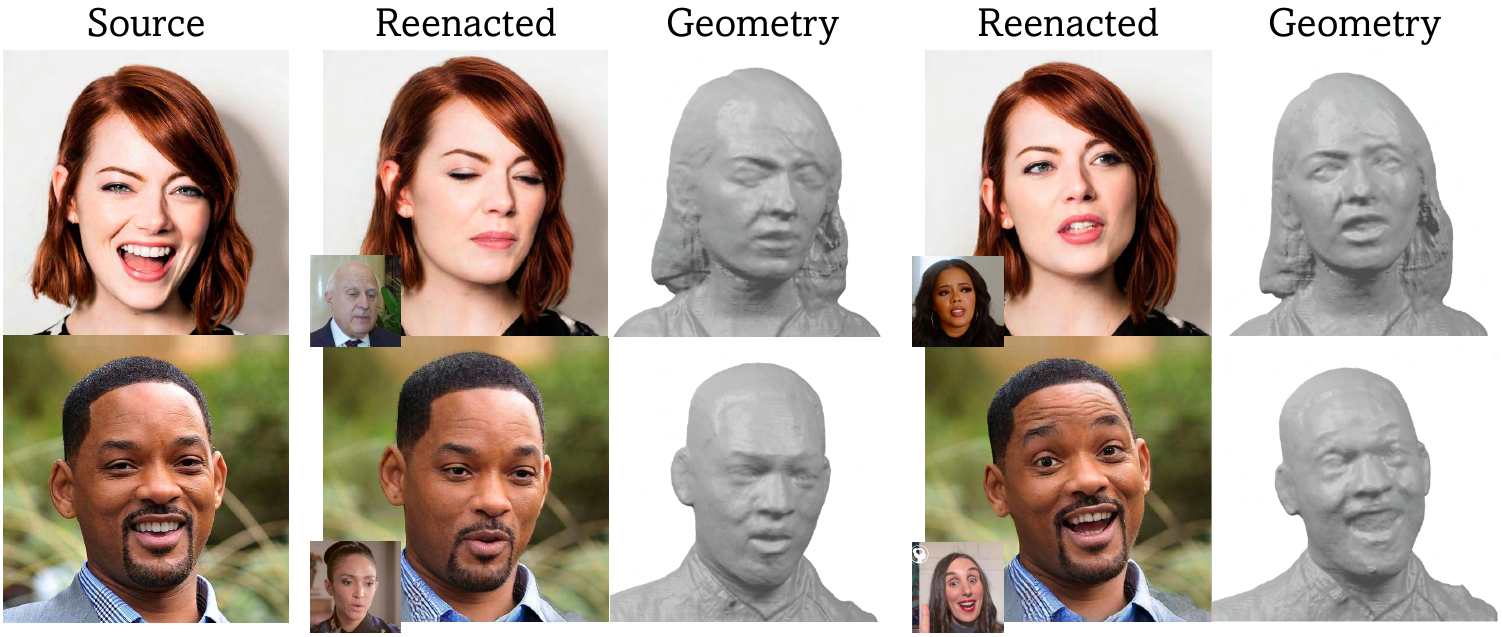}
	\caption{Our method utilizes a feed-forward 4D head synthesizer to create photorealistic head avatars from a single source image. The facial expressions and neck pose of the 3D heads can be precisely controlled by another driving frame (\eg, see the mouth, eye gaze, and forehead wrinkles). The synthesized results also support free-view rendering, thanks to the underlying accurate head geometries. \textbf{Best viewed with zoom-in.} } 
	\label{fig:teaser}
\end{figure*}

This paper suggests a more practical way for learning a one-shot 4D head synthesizer using \emph{pseudo multi-view videos}. The key intuition is to turn existing monocular videos into multi-view ones by learning a static head novel view synthesizer (\ie 3D synthesizer) in advance. This brings us several advantages: \textbf{(1)} Monocular videos exhibit high scalability for learning a generalizable head synthesizer thanks to an easy access to public datasets~\cite{nagrani2017voxceleb,zhu2022celebv,xie2022vfhq} and Internet videos. \textbf{(2)} Monocular head novel view synthesis has been proven achievable through knowledge distillation from a 3D GAN learned solely with in-the-wild images~\cite{trevithick2023real}, thereby enabling the creation of multi-view videos from monocular inputs with minimal effort. \textbf{(3)}~Multi-view videos would facilitate geometry learning and enable using motion representation learned from expressive data beyond 3DMM description. These factors help us avoid a heavy reliance on 3DMM for geometry reconstruction~\cite{khakhulin2022realistic,chu2024gpavatar}, expression manipulation~\cite{li2023one,ye2024real3d,li2023generalizable}, or pose estimation~\cite{li2023one,ma2023otavatar,yu2023nofa}, leading to a faithful 4D head synthesizer with accurate geometry prediction and vivid motion control (see Fig.~\ref{fig:teaser}).

Specifically, we adopt a vision-transformer-based encoder-decoder architecture for the head synthesizer following Portrait4D~\cite{deng2023portrait} to predict triplane-based head NeRF~\cite{chan2021efficient,mildenhall2020nerf} from a source image. A motion embedding~\cite{wang2023progressive} extracted from a driving image is responsible for controlling expression of the predicted tri-planes via motion-related cross-attention layers. At the first stage, we disable all motion-related layers and enforce the synthesizer to learn static tri-plane reconstruction from source images using synthetic data from a pre-trained 3D GAN~\cite{deng2023portrait}. This yields a powerful 3D head synthesizer that can well generalize to real data. At the second stage, we start from the 3D synthesizer and activate all the motion-related components to learn the full model using real videos. We maintain a fixed copy of the pre-trained 3D synthesizer and use it to transform monocular video frames into multi-view ones for cross-view self-reenactment, where we extract motion embedding from a driving frame at an arbitrary viewpoint and learn to reconstruct the same frame from a different random perspective. This way, the geometry priors of the 3D synthesizer can be seamlessly distilled into the full 4D head synthesizer meanwhile detailed motion control can be learned via the self-reenacting process on real videos. Considering that the 3D synthesizer is imperfect, we focus more on reconstructing the original real driving frames and treat any other synthetic multi-view frames as regularization.

It is worth emphasizing that our intention is not to design an advanced network architecture but to explore \textbf{\textit{a novel learning paradigm}} suitable for one-shot 4D head synthesis by creating pseudo multi-view videos from existing monocular ones. This yields a fundamental difference between our method and the previous ones that learn from reconstructing monocular video frames~\cite{li2023one,yu2023nofa,tran2023voodoo,ye2024real3d,chu2024gpavatar}. Certainly, our learning paradigm can be incorporated into these frameworks as well. We demonstrate that by simply using a transformer backbone following~\cite{deng2023portrait}, our method (\ie, Portrait4D-v2) largely outperforms previous approaches in terms of reconstruction fidelity, geometry consistency, and motion control accuracy. We hope that our method will serve as inspiration for future works, encouraging a more seamless integration of 3D priors with in-the-wild 2D data for creating generalizable 4D head avatar synthesizers.

\section{Related Work}
% Photorealistic head avatar creation is a longstanding task and can be dated back to decades ago where Blanz and Vetter created the first 3DMM~\cite{blanz1999morphable} from face scans. Since then, enormous progresses have been made and recent methods~\cite{karras2019style,wang2021one,park2021nerfies,chan2021efficient} enable learning head synthesis models at a much higher fidelity from only images and videos. 
We review approaches falling in the scenario of one-shot head avatar synthesis, which perform subject-agnostic head creation and animation given a single appearance image. 
% Compared to alternative person-specific methods~\cite{kim2018deep,lombardi2021mixture,kirschstein2023nersemble,wang2023styleavatar} that learn to overfit a certain subject with monocular or multi-view videos, the one-shot approaches are more efficient and suitable for large-scale inference. 
We categorize them into 2D-based and 3D-aware ones, depending on whether or not they leverage an explicit camera model for rendering.  

\paragraph{\textbf{2D-based talking head generation.}} The success of CNNs in image generation~\cite{isola2017image,karras2020analyzing} have given rise to plenty of approaches leveraging 2D networks for direct head image synthesis~\cite{burkov2020neural,wang2021one,he2023gaia}. A common strategy among them is to interwine the latent features of appearance and motion within the 2D generative network to achieve photorealisitic and animatable image generation. An earlier work~\cite{zakharov2019few} injects appearance embedding into a U-Net backbone via AdaIN~\cite{huang2017arbitrary} to transfer driving landmarks to reenacted images. \cite{burkov2020neural} and follow-up works~\cite{zhou2021pose,wang2023progressive} learn disentangled latent representations of appearance and motion as input to a StyleGAN-like~\cite{karras2020analyzing} generator. A prevailing trend in recent methods~\cite{siarohin2019first, wang2021one, ren2021pirenderer,hong2022depth,drobyshev2022megaportraits,zhang2023metaportrait} involves enforcing a warping field onto intermediate appearance feature maps, leveraging the inherent characteristics of facial movements. However, these methods fall short in maintaining the head geometry consistency under large pose changes due to a lack of 3D understanding. While some methods~\cite{wang2021one, drobyshev2022megaportraits} introduce 3D inductive bias into the framework, they cannot fully alleviate the inconsistency issue with their learned non-physical projection process from 3D to 2D. In addition, none of the above methods support free-view rendering owing to the absence of an explicit camera model.

\paragraph{\textbf{3D-aware head avatar synthesis.}} To allow free-view rendering with more strict 3D consistency, recent works incorporate 3D representations and perspective camera projection into the head image synthesis process. Earlier methods~\cite{xu2020deep,khakhulin2022realistic} utilize morphable meshes to represent head geometries and textures, and successive works~\cite{hong2022headnerf,ma2023otavatar,li2023one,yu2023nofa,li2023generalizable,chu2024gpavatar,ye2024real3d} leverage more advanced representations such as NeRFs~\cite{mildenhall2020nerf,chan2021efficient} to better model hairs and other accessories. Notably, introducing the 3D representations poses significant challenges for the model to learn monocular 3D reconstruction and animation from 2D videos, which is highly ill-posed without leveraging 3D priors or multi-view data. Consequently, some methods~\cite{hong2022headnerf,zhuang2022mofanerf} resort to multi-view videos collected from constrained environment as supervision, yet the limited diversity of the training data restricts their generalizability to unconstrained images. Another line of works stick to in-the-wild videos as supervision and leverage monocular 3DMM reconstructors~\cite{deng2019accurate,feng2021learning,danvevcek2022emoca} to facilitate 3D geometry and motion learning, by using reconstructed 3DMM meshes for head modeling~\cite{xu2020deep,khakhulin2022realistic}, estimating camera poses for feature transformation~\cite{li2023one} and image rendering~\cite{ma2023otavatar,yu2023nofa}, or utilizing 3DMM expressions for animation synthesis~\cite{li2023one,li2023generalizable,chu2024gpavatar,ye2024real3d}. Nevertheless, the limited reconstruction accuracy and expressiveness of 3DMM hinder them from synthesizing faithful head avatars. \cite{tran2023voodoo} avoids using 3DMM and achieves expression disentanglement in a data-driven manner leveraging a 3D head synthesizer~\cite{trevithick2023real} for pose canonicalization. Still, it uses monocular videos as supervision which differs from our learning paradigm. A recent method Portrait4D~\cite{deng2023portrait} learns from synthetic data of multi views. However, its data is generated by a 4D GAN with 3DMM-based expression control, which exhibits a non-negligible domain gap when compared to real videos of subtle facial expressions, and thus can limit the generalizability of the learned model. 
% Therefore, it experiences inferior head reenactment fidelity compared to our method that learns from real videos.

\section{Preliminaries: Triplane-Based 3D Representation}
\label{sec:triplane}
We follow previous successes~\cite{chan2021efficient,li2023one,trevithick2023real,deng2023portrait} to model 3D head via a triplane-based NeRF representation~\cite{chan2021efficient}. Tri-plane depicts a 3D volumetric space using three orthogonal 2D feature planes $\mathbf{T}_{xy},\mathbf{T}_{yz},\mathbf{T}_{zx}$, each representing projected 3D features onto corresponding axis-aligned planes $xy,yz,zx$, respectively. Given a point $\mathbf{x}\in \mathbb{R}^3$, its volume density $\sigma \in \mathbb{R}$ and color $\mathbf{c} \in \mathbb{R}^{d_c}$ can be decoded from the tri-plane via:
\begin{equation}
\label{eq:triplane}
(\mathbf{c}, \sigma) = \mathrm{MLP}(\mathbf{T}_{xy}(\mathbf{x})+\mathbf{T}_{yz}(\mathbf{x})+\mathbf{T}_{zx}(\mathbf{x})),
\end{equation}
where $\mathbf{T}_{\cdot}(\mathbf{x})$ denotes projecting $\mathbf{x}$ onto the corresponding plane for feature acquisition, and $\mathrm{MLP}$ decodes the summed feature to the final radiance (\ie, color and density). Following volume rendering process~\cite{kajiya1984ray} as in NeRF~\cite{mildenhall2020nerf}, one can obtain a 2D feature image $I_{c}$ (usually the first three channels represent RGB color) from the 3D space by accumulating radiance of each point along rays from a given camera viewpoint $\bm \theta$. The rendered image $I_{c}$ is then sent into a 2D super-resolution module~\cite{chan2021efficient, wang2021towards} yielding a final synthesized image $I$ at a higher resolution. For brevity, we use a single renderer $\mathcal{R}$ to represent the above rendering process from the tri-plane $\mathbf{T}$ to the final image $I$, leading to $I=\mathcal{R}(\mathbf{T}, \bm \theta)$.

\section{Method}

Our goal is to synthesize an animatable 3D head represented by the tri-plane $\mathbf{T}$ from a source image $\hat{I}_{s}$, where $\mathbf{T}$ faithfully reconstructs the source appearance and its head motion mimics that from another driving image $\hat{I}_{d}$. We adopt the feed-forward backbone of Portrait4D~\cite{deng2023portrait} which directly predicts $\mathbf{T}$ from the input images via a transformer-based reconstructor $\mathrm{\Psi}$ (Sec.~\ref{sec:reconstructor}). To learn $\mathrm{\Psi}$ for faithful head avatar synthesis, we create pseudo multi-view videos from monocular ones as supervision, by first learning a 3D head synthesizer $\mathrm{\Psi}_{3d}$ to generate novel views of each video frame (Sec.~\ref{sec:3d}), and then leveraging the synthesized multi-view videos to perform cross-view self-reenactment for training $\mathrm{\Psi}$ (Sec.~\ref{sec:4d}). Figure~\ref{fig:framework} shows the overview and we describe each part in detail below.

\subsection{Revisiting Portrait4D Reconstructor}
\label{sec:reconstructor}
\begin{figure*}[t]
	\small
	\centering
	\includegraphics[width=0.98\textwidth]{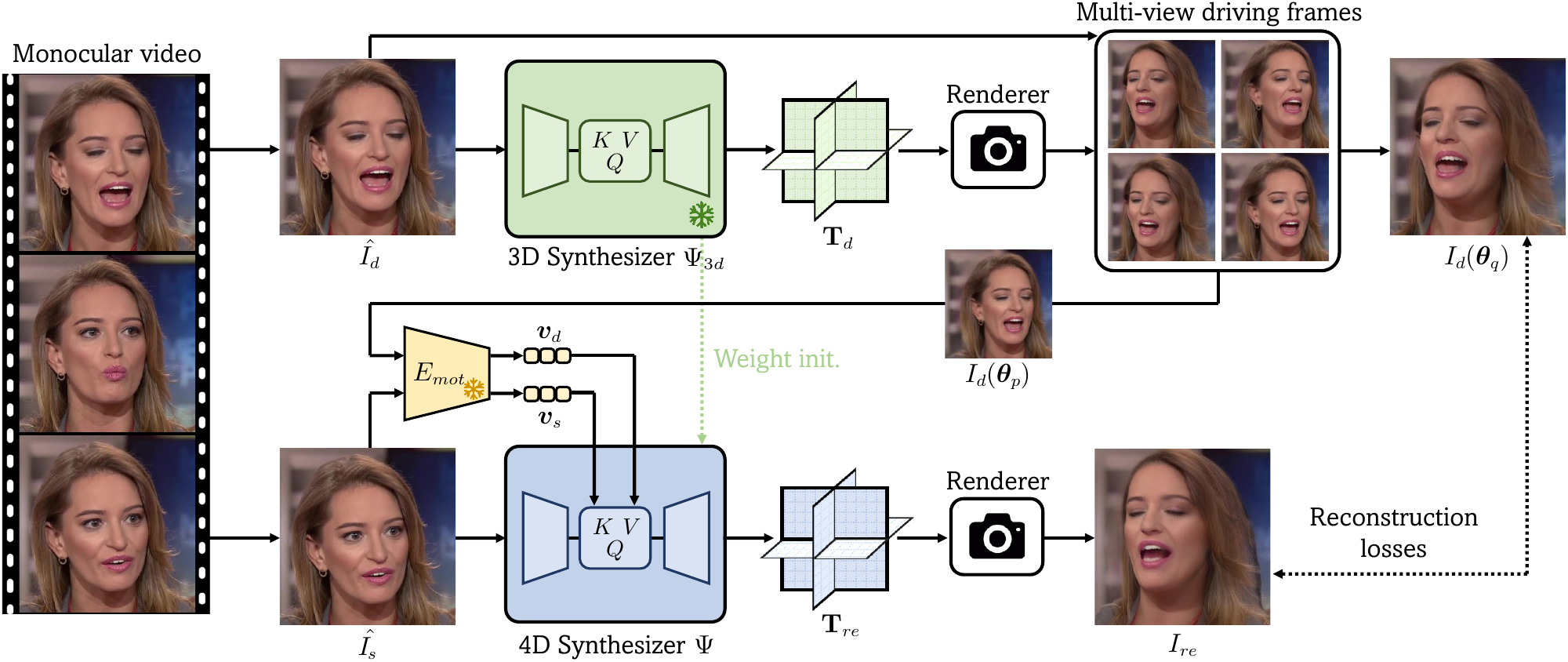}
	\caption{Overview of our approach. Given a monocular training video, we first leverage a pre-trained 3D synthesizer $\mathrm{\Psi}_{3d}$ to turn each driving frame within the video into multi-view one, and then use the pseudo multi-view driving frames and a source frame sampled from the original video to perform cross-view self-reenactment for learning a feed-forward 4D head synthesizer $\mathrm{\Psi}$. After training, $\mathrm{\Psi}$ can synthesize an animatable 3D head given two arbitrary images as the source and driving, respectively.}
	\label{fig:framework}
\end{figure*}

We begin by revisiting the tri-plane reconstructor $\mathrm{\Psi}$ of Portrait4D which uses a hybrid structure of CNNs and Vision Transformer (ViT) blocks. Specifically, two CNN encoders are first leveraged to extract global and detail appearance feature maps from the source image, respectively. The global feature map is then sent to several ViT blocks for pose canonicalization, and further concatenated with the detail feature map and fed into a ViT decoder with convolutions to generate the final tri-plane $\mathbf{T}$. To inject motion information for reenactment, Portrait4D introduces additional cross-attention layers into each canonicalization ViT block, where the cross-attentions in the first several blocks receive motion embedding $\bm v_s$ of the source image for expression neutralization and the remaining receive that of the driving image (\ie, $\bm v_d$) for reenactment (see Fig.~\ref{fig:framework}). For faithful expression reconstruction, Portrait4D leverages a pre-trained 2D motion encoder $E_{mot}$~\cite{wang2023progressive} learned from video reenactment to extract the motion embedding.

\paragraph{\textbf{Characteristics of the reconstructor.}} An advantage of $\mathrm{\Psi}$ is that it is purely data-driven without being restricted by 3DMM-related inductive bias as in many previous approaches~\cite{li2023one,yu2023nofa,ye2024real3d}. However, it also necessitates extensive and precise 3D data for effective training of the 4D head synthesizer. Otherwise, the network is inclined to synthesize 3D head with flattened geometries (see Fig.~\ref{fig:ablation}). Portrait4D addresses this challenge by training on 4D synthetic data generated by a pre-trained GAN~\cite{deng2023portrait}, yet the synthesized expressions remain confined to the 3DMM description space which cannot fully exploit the expressive capability of the motion embedding. Besides, the synthetic identities have domain gaps with the real ones which can be detrimental to the model's generalizability.

\begin{figure*}[t]
	\small
	\centering
 \begin{subfigure}[b]{0.48\textwidth}
 \centering
\includegraphics[width=0.96\textwidth]{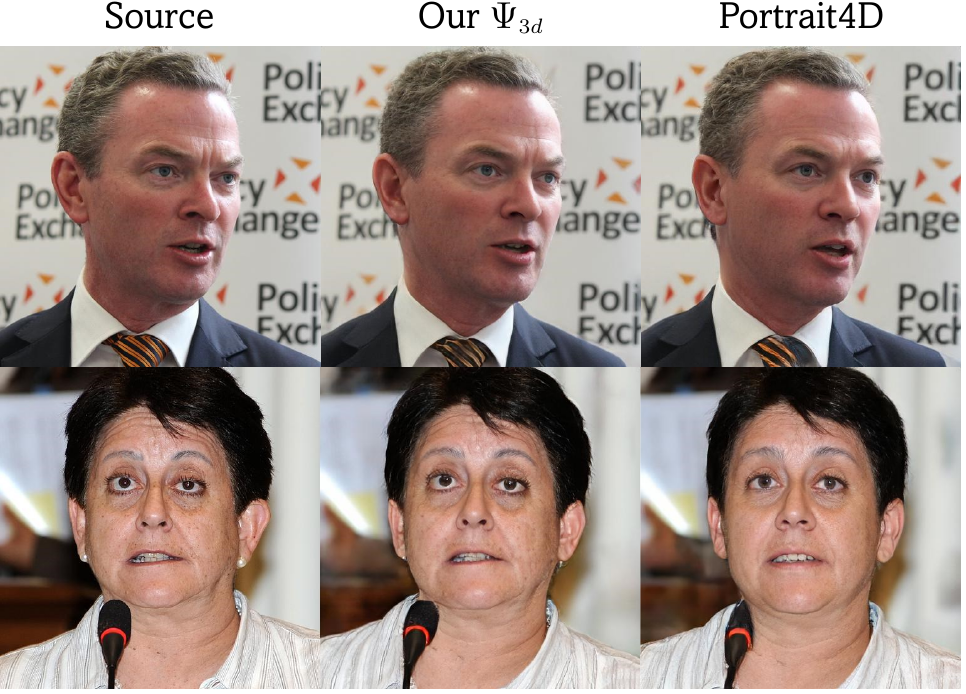}
	\caption{}
	\label{fig:static_compare}     
 \end{subfigure}
 \hfill
  \begin{subfigure}[b]{0.48\textwidth}
\centering
\includegraphics[width=0.96\textwidth]{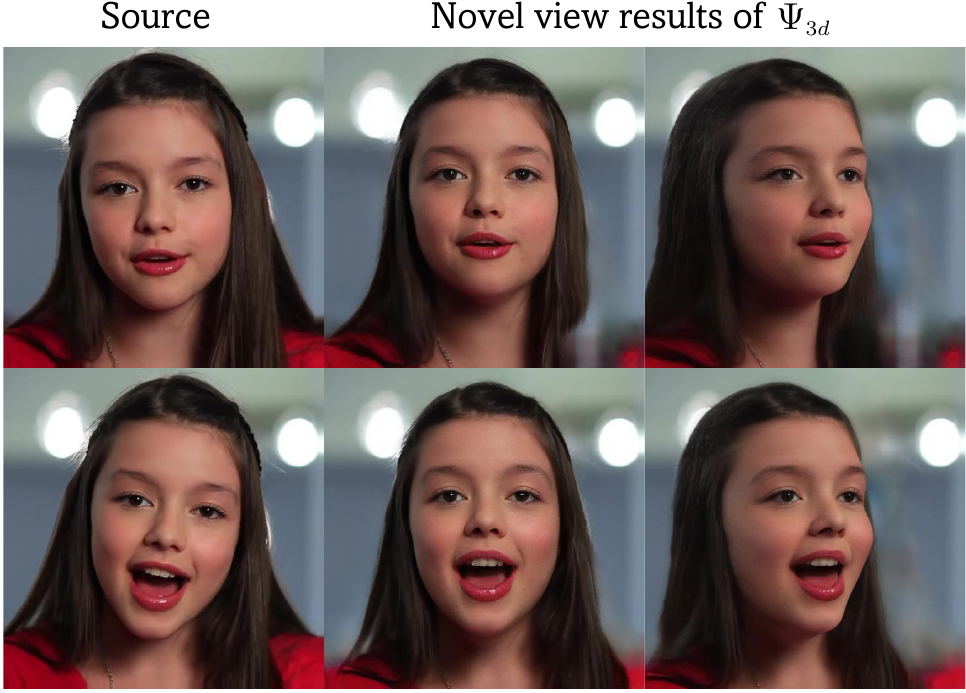}
	\caption{}
	\label{fig:static_mv}     
 \end{subfigure}
 \caption{(a) Self-reconstruction comparison between $\mathrm{\Psi}_{3d}$ and \cite{deng2023portrait} learned with the same data.  $\mathrm{\Psi}_{3d}$ yields better reconstruction fidelity. (b)~$\mathrm{\Psi}_{3d}$ learned using static images is capable of maintaining geometry consistency across different frames within a video clip.}
 \label{fig:static}
\end{figure*}

An interesting observation is that if we remove the motion-related cross-attentions in $\mathrm{\Psi}$ and use the same synthetic data from Portrait4D to train a static 3D head synthesizer\footnote{Under this circumstance, the problem degenerates to that of~\cite{trevithick2023real} for head novel view synthesis, which has been proven achievable using synthetic data from EG3D~\cite{chan2021efficient}.}, it can well generalize to real images with nuanced expressions hard to be described by 3DMM, as shown in Fig.~\ref{fig:static_compare}. This inspires us that instead of generating 4D data directly from scratch as supervision, it is more practical to learn a 3D head synthesizer in advance and use it to create multi-view videos from existing monocular ones for training, as described below.

\subsection{3D Synthesizer for Multi-View Video Creation}
\label{sec:3d}
We learn a 3D head synthesizer to turn an arbitrary frame to its corresponding tri-plane for free-view rendering. Following Sec.~\ref{sec:reconstructor}, we adopt the reconstructor $\mathrm{\Psi}$ as backbone with all its cross-attention layers disabled, denoted as $\mathrm{\Psi}_{3d}$. While previous methods~\cite{chan2021efficient,ye2024real3d,tran2023voodoo} have learned similar 3D head synthesizers using synthetic data from EG3D~\cite{chan2021efficient}, we do not follow them but leverage the data generated by the 4D GAN of Portrait4D (namely GenHead~\cite{deng2023portrait}) as supervision. Compared to EG3D, GenHead separates foreground head with background image, and its generated data are aligned with the scale of FLAME~\cite{li2017learning}, allowing further neck pose rotation via an $SO(3)$ field derived from the FLAME mesh.

\paragraph{\textbf{Learning the 3D synthesizer.}} Specifically, GenHead requires FLAME parameters as input condition for data synthesis. Therefore, we first apply off-the-shelf 3DMM reconstructors~\cite{deng2019accurate,bfmtoflame} with further landmark-based optimization to in-the-wild images to extract FLAME parameters $(\bm \alpha, \bm \beta, \bm \gamma)$ from them, where $\bm \alpha \in \mathbb{R}^{300}$ is a shape code, $\bm \beta \in \mathbb{R}^{100}$ is an expression code, and $\bm \gamma \in \mathbb{R}^{9}$ is a pose code. 
% This leads to a sample set $\mathcal{S}=\{\bm \alpha_i \}\times\{\bm \beta_i \}\times\{\bm \gamma_i \}$ where the subscription $i$ denotes the parameters from the $i$'s image. 
We randomly sample $(\bm \alpha, \bm \beta, \bm \gamma)$ triplet among the extracted parameter set and combine it with a random Gaussian noise $\bm z \in \mathbb{R}^{512}$ as input to GenHead to create arbitrary 3D heads. We then sample camera extrinsics $\bm \theta \in SE(3)$ from a pre-defined uniform distribution and render each generated 3D head to its multi-view images $\{\bar{I}(\bm \theta_i)\}$. To learn the 3D synthesizer $\mathrm{\Psi}_{3d}$, we send an arbitrary synthetic image $\bar{I}(\bm \theta_p)$ of a 3D head to $\mathrm{\Psi}_{3d}$, and enforce it to reconstruct a tri-plane $\mathbf{T}$ of the input that is consistent with another image $\bar{I}(\bm \theta_q)$ of the 3D head when rendered at an arbitrary view $\bm \theta_q$. Following~\cite{deng2023portrait}, we apply multi-level supervisions to $\mathrm{\Psi}_{3d}$ to ensure its faithful reconstruction of the synthetic data:
\begin{equation}
\mathcal{L} = \mathcal{L}_{1} + \mathcal{L}_{\mathrm{LPIPS}} + \mathcal{L}_{id} + \mathcal{L}_{adv} + \mathcal{L}_{depth} + \mathcal{L}_{opa} + \mathcal{L}_{T}, \label{eq:training_3d}
\end{equation}
where $\mathcal{L}_{1}$ is the L1 distance between reconstructed images of $\mathrm{\Psi}_{3d}$ and the synthetic images of GenHead, $\mathcal{L}_{\mathrm{LPIPS}}$ is the perceptual difference~\cite{zhang2018unreasonable} between them, $\mathcal{L}_{id}$ calculates negative cosine similarity between face recognition features~\cite{deng2019arcface} of the images, $\mathcal{L}_{adv}$ is the adversarial loss of~\cite{deng2023portrait}, $\mathcal{L}_{depth}$ and $\mathcal{L}_{opa}$ calculate L1 difference of rendered depth and opacity images between $\mathrm{\Psi}_{3d}$'s reconstructions and the synthetic data, respectively, and $\mathcal{L}_{T}$ computes the L1 distance of sampled tri-plane features between the reconstructions and the ground truth.

\paragraph{\textbf{Synthesizing multi-view videos.}} We use the learned $\mathrm{\Psi}_{3d}$ to turn a monocular video $\hat{V}=\{\hat{I}_t\}_{t=1:T}$ to its multi-view version $V_{mv}=\{\{I_t(\bm \theta_i)\}_{i=1:N}\}_{t=1:T}$ via
\begin{equation}
 I_t(\bm \theta_i) = \mathcal{R}(\mathbf{T}_t, \bm \theta_i) = \mathcal{R}(\mathrm{\Psi}_{3d}(\hat{I}_t), \bm \theta_i), \label{eq:mv}
\end{equation}
where $\mathcal{R}$ is the renderer defined in Sec.~\ref{sec:triplane}.
Although $\mathrm{\Psi}_{3d}$ is learned using static multi-view images, we empirically find that it can maintain the geometry consistency between different frames in a video clip to some extent (see Fig.~\ref{fig:static_mv}). What's more, it better preserves detailed shape and expression of a real frame compared to Portrait4D (as in Fig.~\ref{fig:static_compare}), even though they are trained with the same synthetic data generated by the 3DMM-conditioned 4D GAN. We conjecture this is due to that: (1) the static reconstruction process utilizes less 3DMM inductive bias within the data as it does not tackle expression change constrained by 3DMM deformtation; (2) the synthetic data have certain degrees of freedom beyond 3DMM thanks to the additional noise $\bm z$. With the pseudo multi-view videos created via Eq.~\eqref{eq:mv}, we learn our final 4D head synthesizer $\mathrm{\Psi}$ as follows.

\subsection{Cross-View Self-Reenactment Learning}
\label{sec:4d}
We learn the 4D synthesizer $\mathrm{\Psi}$ via cross-view self-reenactment, leveraging real videos $\hat{V}$ and their corresponding multi-view versions $V_{mv}$. As shown in Fig.~\ref{fig:framework}, during each iteration, we randomly select a source frame $\hat{I}_s$ from $\hat{V}$ and two free-view driving images $I_d(\bm \theta_p)$ and $I_d(\bm \theta_q)$ of the same motion from $V_{mv}$. We let the synthesizer $\mathrm{\Psi}$ to take $\hat{I}_s$ and $I_d(\bm \theta_p)$ as input and synthesize a reenacted image $I_{re}$ at $I_d(\bm \theta_q)$'s camera viewpoint $\bm \theta_q$:
\begin{equation}
 I_{re} = \mathcal{R}(\mathbf{T}, \bm \theta_q) = \mathcal{R}(\mathrm{\Psi}(\hat{I}_s, \bm v_s, \bm v_d), \bm \theta_q), \label{eq:4d}
\end{equation}
where $\bm v_s$ and $\bm v_d$ are the motion embeddings of $\hat{I}_s$ and $I_d(\bm \theta_p)$ extracted via the motion encoder $E_{mot}$~\cite{wang2023progressive}, respectively. We impose $I_{re}$ to match the content of $I_d(\bm \theta_q)$ through a set of losses:

\begin{equation}
\mathcal{L} = \mathcal{L}_{1} + \mathcal{L}_{\mathrm{LPIPS}} + \mathcal{L}_{id} + \mathcal{L}_{adv}, \label{eq:training_4d}
\end{equation}
where the four terms of $\mathcal{L}$ are similarly defined in Eq.~\eqref{eq:training_3d} and calculated between $I_{re}$ and $I_d(\bm \theta_q)$. We initialize $\mathrm{\Psi}$ using the pre-trained weights of $\mathrm{\Psi}_{3d}$ except the motion-related components which are initialized randomly. 

Considering that $I_d(\bm \theta_p)$ and $I_d(\bm \theta_q)$ are generated by an imperfect 3D synthesizer, we replace either of them with the original driving frame $\hat{I}_d$ from $V$ with certain possibilities to let the 4D synthesizer $\mathrm{\Psi}$ be supervised by reenacting the real videos as well.
% \footnote{If $\hat{I}(d)$ replaces $I_d(\bm \theta_q)$ for loss computation, we render the reenacted image $I_{re}$ at corresponding camera pose $\bm \theta_d$ of $\hat{I}_d$ where $\bm \theta_d$ is obtained using the 3DMM reconstruction process described in Sec.~\ref{sec:3d}.}.
We substitute $I_d(\bm \theta_p)$ with a probability of $10\%$ and $I_d(\bm \theta_q)$ with a probability of $80\%$. Therefore, $\mathrm{\Psi}$ concentrates more on reconstructing the original real frame $\hat{I}_d$ while the pseudo free-view frame $I_d(\bm \theta_q)$ serves more like a geometry regularization. We apply a lower substitution rate for $I_d(\bm \theta_p)$ and a higher one for $I_d(\bm \theta_q)$, this is because the former is only used for extracting high-level motion embedding where the imperfect result has less influence compared to that of the latter used for supervising pixel-wise image reconstruction. Besides, we use all the four losses in Eq.~\eqref{eq:training_4d}  if $\hat{I}_d$ is used as the ground truth and otherwise (\ie, $I_d(\bm \theta_q)$ is used) only $\mathcal{L}_{\mathrm{LPIPS}}$. 

\paragraph{\textbf{Effect of cross-view learning.}} Our learning process distills the geometry and appearance priors within the 3D synthesizer $\mathrm{\Psi}_{3d}$ into the 4D synthesizer $\mathrm{\Psi}$, as the latter is enforced to mimic the reconstruction of an arbitrary image $I_d(\bm \theta_q)$ by the former at any camera pose $\bm \theta_q$. This yields more plausible reconstruction result for unseen regions compared to learning with only monocular supervision, as we will show in Sec.~\ref{sec:ablation}. What's more, our strategy enables using a non-3D-aware motion embedding learned from 2D reenactment~\cite{wang2023progressive}. By leverage $I_d(\bm \theta_p)$ at an arbitrary view $\bm \theta_p$ to extract the motion embedding $\bm v_d$ for reenactment, any pose-related information within $\bm v_d$ can be squeezed out. As a result, $\mathrm{\Psi}$ can well utilize the detailed expressive ability of the 2D motion embedding meanwhile ignore its pose-related information to achieve strong 3D consistency.
% Otherwise, the 4D synthesizer $\mathrm{\Psi}$ cannot faithfully reconstruct $I_d(\bm \theta_q)$ and $\hat{I}_d$ when their poses are unrelated to that of $I_d(\bm \theta_p)$. 
% As a result, it is possible to leverage advanced motion encoders~\cite{wang2023progressive,he2023gaia} learned from 2D reenactment for our motion control.

\begin{figure*}[t]
	\small
	\centering
	\includegraphics[width=0.97\textwidth]{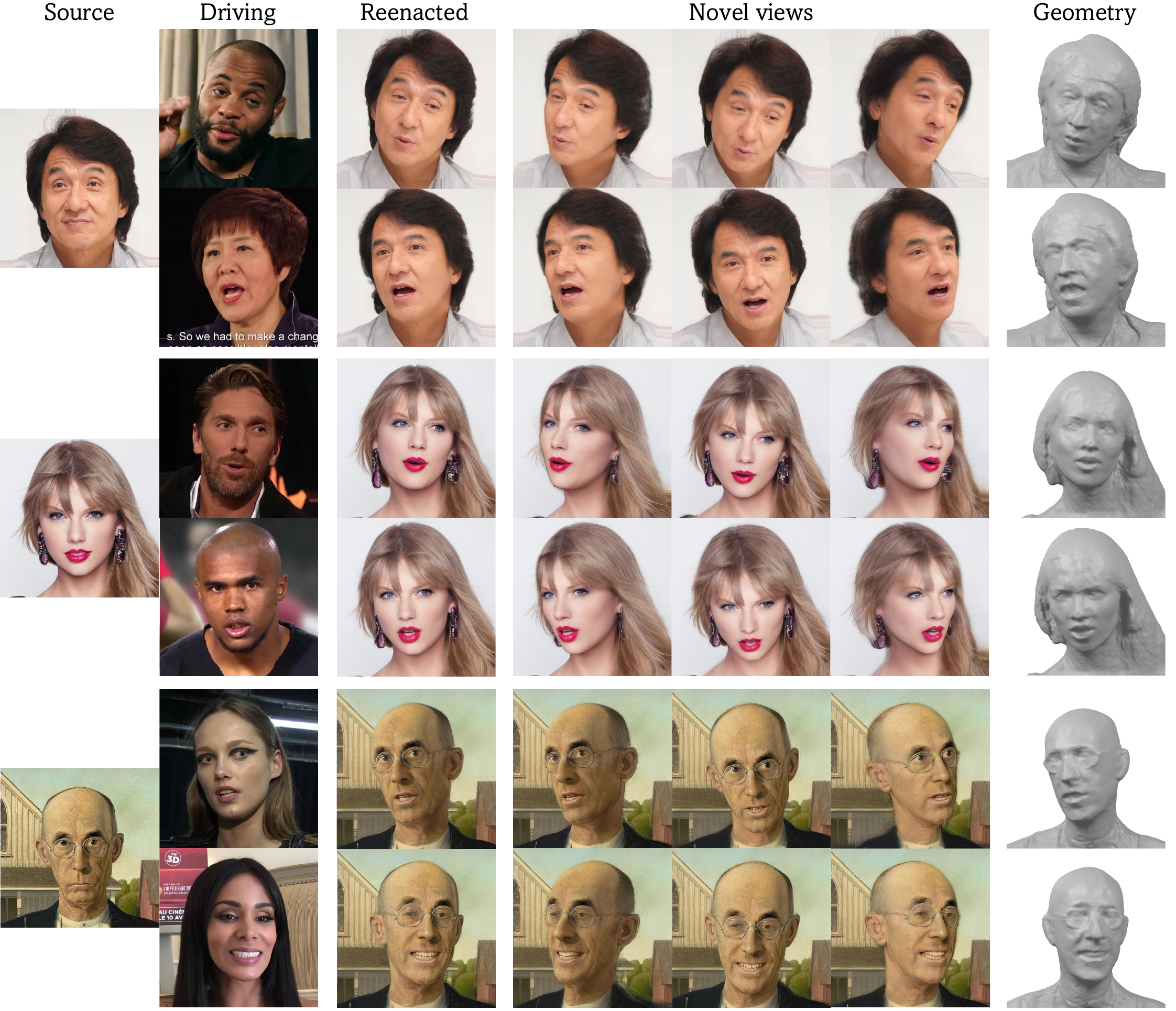}
	\caption{One-shot head synthesis results of our method on in-the-wild images.}
	\label{fig:main_result}
\end{figure*}

\section{Experiment}
% \subsection{Implementation Details}
\paragraph{\textbf{Implementation.}} To learn $\mathrm{\Psi}_{3d}$, we use GenHead~\cite{deng2023portrait} trained on FFHQ~\cite{karras2019style} at $512^2$ to synthesize multi-view supervisions. 
% The FLAME conditions for image synthesis are extracted from FFHQ as well. For camera pose, we follow~\cite{deng2023portrait} to sample it from a pre-defined uniform distribution.
% with pitch from $[-0.25,0.65]$ $rad$, yaw from $[-0.78,0.78]$ $rad$, and roll from $[-0.25,0.25]$ $rad$. We also sample the camera radius from $[3.65,4.45]$, and the camera look-at position from $[-0.01,0.01]\times[-0.01,0.01]\times[0.02,0.04]$. During rendering, the camera intrinsics are fixed with a $12^\circ$ field of view (fov). 
We adopt an Adam optimizer~\cite{kingma2015adam} with a learning rate of $1e-4$ and a batch size of $32$, and train $\mathrm{\Psi}_{3d}$ to see $10$M images in total. 
% which takes 8 days on 8 Tesla A100 GPUs with 80GB memory.
Then, we learn the 4D synthesizer $\mathrm{\Psi}$ using a training split of $10$K video clips from VFHQ~\cite{xie2022vfhq}. 
% and sample $50$ frames within each video clip. 
% During each iteration, we randomly choose $2$ frames in a video clip as the source and driving, respectively, and send the driving image to the pre-trained $\mathrm{\Psi}_{3d}$ to obtain its pseudo multi-view frames. 
% where we use a similar camera distribution for rendering as described above. 
The model is trained to see $6$M images, using Adam with a learning rate of $2.5e-4$ for the motion-related layers and $1e-4$ for the remaining, and a batch size of $32$. 
% which also takes 8 days on the same GPU configuration. 
See Sec.~\ref{sec:implement} for more details.

\subsection{One-Shot Head Synthesis Results}\label{sec:result}
Figure~\ref{fig:teaser} and~\ref{fig:main_result} demonstrate our one-shot head reenactment results on in-the-wild images. Our method faithfully reconstructs the source appearance meanwhile mimics the nuanced expressions in different driving images. It can also model the dynamic wrinkles brought by exaggerated facial motions. What's more, our method supports free view rendering of the reconstructed heads thanks to the underlying tri-plane representation. The accurately reconstructed head geometries (visualized via Marching Cubes~\cite{lorensen1987marching}) guarantee strong 3D consistency across different views. While our method is trained on real videos, it can also handle artistic portraits as depicted by the last case in Fig.~\ref{fig:main_result}.

\paragraph{\textbf{Inference speed.}} Our method runs at $10$ FPS with a batch size of $1$ on an A100 GPU using the naive Pytorch framework without specialized acceleration. Further caching the neutralized feature map of the source image leads to $15$ FPS.

%%%%%%%%%%%%%%%%%%%%%% remove LPIPS for cross-id

\begin{table*}[t]
\centering
\caption{Self- and cross-ID reenactment results on VFHQ at $512^2$. LPIPS$_h$ is for head region only. \tikzcircle[gold,fill=gold]{2pt}, \tikzcircle[silver,fill=silver]{2pt}, and \tikzcircle[bronze,fill=bronze]{2pt} denote the 1st, 2nd and 3rd places, respectively. }
% \resizebox{\textwidth}{20mm}{
\begin{adjustbox}{width=0.8\columnwidth,center}
\begin{tabular}{l|llllll|llll}
\toprule

\multicolumn{1}{c}{\multirow{2}{*}{Method}} &  \multicolumn{6}{c}{Self reenactment}  & \multicolumn{4}{c}{Cross-ID reenactment} \\ 
\cline{2-11} 
 \multicolumn{1}{c}{} & LPIPS$_{h}$$\downarrow$ & LPIPS $\downarrow$ &  FID $\downarrow$ & ID $\uparrow$ & AED $\downarrow$  & APD $\downarrow$ & FID $\downarrow$ & ID $\uparrow$ & AED $\downarrow$ & APD $\downarrow$\\
 \midrule
 PIRenderer~\cite{ren2021pirenderer} & 0.213 & 0.323 &  56.2 & 0.801  & 0.034  & 1.402 &  78.6 & 0.582 & 0.146\tikzcircle[gold,fill=gold]{2pt} &  2.833\\ 
Face-vid2vid~\cite{wang2021one} & 0.187 & 0.289\tikzcircle[bronze,fill=bronze]{2pt} & 52.0 & 0.826\tikzcircle[silver,fill=silver]{2pt} & 0.028\tikzcircle[silver,fill=silver]{2pt} & 1.603 & 79.2 & 0.606 & 0.178 &  5.188 \\
StyleHEAT~\cite{yin2022styleheat} & 0.224 & 0.320 & 66.9 & 0.633  & 0.053  &  1.543 & 81.7 & 0.499 & 0.165 & 5.864 \\
\midrule
ROME~\cite{khakhulin2022realistic} & 0.326 & 0.594 & 99.8 & 0.660 & 0.036 & 1.007 & 108 & 0.532 & 0.148\tikzcircle[silver,fill=silver]{2pt} &  2.294\\
OTAvatar~\cite{ma2023otavatar} & 0.291 & 0.556 & 110  &  0.436 & 0.080 & 3.650 &107& 0.350 & 0.194 & 5.779 \\
HideNeRF~\cite{li2023one} & 0.282  & 0.417 & 74.5 & 0.800 & 0.035 &  2.182 & 105&0.491&0.166 & 4.242 \\
Real3DPortrait~\cite{ye2024real3d} & 0.190  & 0.283\tikzcircle[silver,fill=silver]{2pt} & 44.6\tikzcircle[bronze,fill=bronze]{2pt} & 0.802\tikzcircle[bronze,fill=bronze]{2pt} & 0.029\tikzcircle[bronze,fill=bronze]{2pt} &  1.180 & 61.3\tikzcircle[bronze,fill=bronze]{2pt}&0.721\tikzcircle[silver,fill=silver]{2pt}& 0.183 & 3.951 \\
GPAvatar~\cite{chu2024gpavatar} & 0.191  & 0.418 & 63.6 & 0.786 & 0.028\tikzcircle[silver,fill=silver]{2pt}  & 0.747 & 80.6&0.570&0.154\tikzcircle[bronze,fill=bronze]{2pt} & 2.409 \\
GOHA~\cite{li2023generalizable} & 0.179\tikzcircle[silver,fill=silver]{2pt}   & 0.373 & 52.0 & 0.677 & 0.048 & 0.697\tikzcircle[bronze,fill=bronze]{2pt} & 68.7 &0.493&0.169 &1.152\tikzcircle[bronze,fill=bronze]{2pt} \\

Portrait4D~\cite{deng2023portrait} & 0.181\tikzcircle[bronze,fill=bronze]{2pt} & 0.320 & 43.0\tikzcircle[silver,fill=silver]{2pt} & 0.773 & 0.033 & 0.612\tikzcircle[silver,fill=silver]{2pt} & 54.8\tikzcircle[silver,fill=silver]{2pt} &0.620\tikzcircle[bronze,fill=bronze]{2pt} & 0.164 &  1.020\tikzcircle[silver,fill=silver]{2pt} \\

\midrule

Ours& 0.139\tikzcircle[gold,fill=gold]{2pt} & 0.224\tikzcircle[gold,fill=gold]{2pt} & 26.3\tikzcircle[gold,fill=gold]{2pt} & 0.873\tikzcircle[gold,fill=gold]{2pt} & 0.019\tikzcircle[gold,fill=gold]{2pt} & 0.580\tikzcircle[gold,fill=gold]{2pt} & 48.5\tikzcircle[gold,fill=gold]{2pt} & 0.736\tikzcircle[gold,fill=gold]{2pt} & 0.161 & 0.858\tikzcircle[gold,fill=gold]{2pt} \\
\bottomrule
\end{tabular}
\end{adjustbox}
% }

\label{tab:main_self}
\end{table*}
%%%%%%%%%%%%%%%%%%%%%% 

\begin{table*}[t]\footnotesize
\centering
\caption{(a) Evaluation on cross-ID expression similarity. (b) A user study, where the participants are asked to choose one best method among all.}
    \begin{subtable}[h]{0.4\textwidth}
        \centering
    \caption{Cross-ID expression accuracy}
    \begin{adjustbox}{width=0.8\columnwidth,center}
        \begin{tabular}{l|cc}
        \toprule
        Method & AED$_f$ $\downarrow$ & SyncNet $\uparrow$ \\
 \midrule
     PIRenderer~\cite{ren2021pirenderer} & 0.209 & 3.11 \\
     ROME~\cite{khakhulin2022realistic} & 0.212 & 2.99 \\
     Real3DPortrait~\cite{ye2024real3d} & 0.252 & 3.49 \\
     GPAvatar~\cite{chu2024gpavatar} & 0.200 & 2.95 \\
     Portrait4D~\cite{deng2023portrait} & 0.205 & 3.23\\
     \hline
     Ours & \textbf{0.185} & \textbf{3.66} \\
     \bottomrule
       \end{tabular}
       \end{adjustbox}
       \label{tab:cross-exp}
    \end{subtable}
    \begin{subtable}[h]{0.5\textwidth}
        \centering
    \caption{User study}
    \begin{adjustbox}{width=0.9\columnwidth,center}
        \begin{tabular}{l|cc}
        \toprule
        Method & Image quality $\uparrow$ & Expression Acc. $\uparrow$ \\
 \midrule
     PIRenderer~\cite{ren2021pirenderer} & 0.47\% & 1.40\% \\
     ROME~\cite{khakhulin2022realistic} & 0.47\% & 3.72\% \\
     Real3DPortrait~\cite{ye2024real3d} & 0.47\% & 2.33\% \\
     GPAvatar~\cite{chu2024gpavatar} & 6.28\% & 14.7\% \\
     Portrait4D~\cite{deng2023portrait} & 11.6\% & 6.28\% \\
     \hline
     Ours & \textbf{80.7\%} & \textbf{71.6\%} \\
     \bottomrule
       \end{tabular}
       \end{adjustbox}
       \label{tab:user-study}
    \end{subtable}
\label{tab:cross-id}
\end{table*}

\begin{figure*}[t]
	\small
	\centering
	\includegraphics[width=0.97\textwidth]{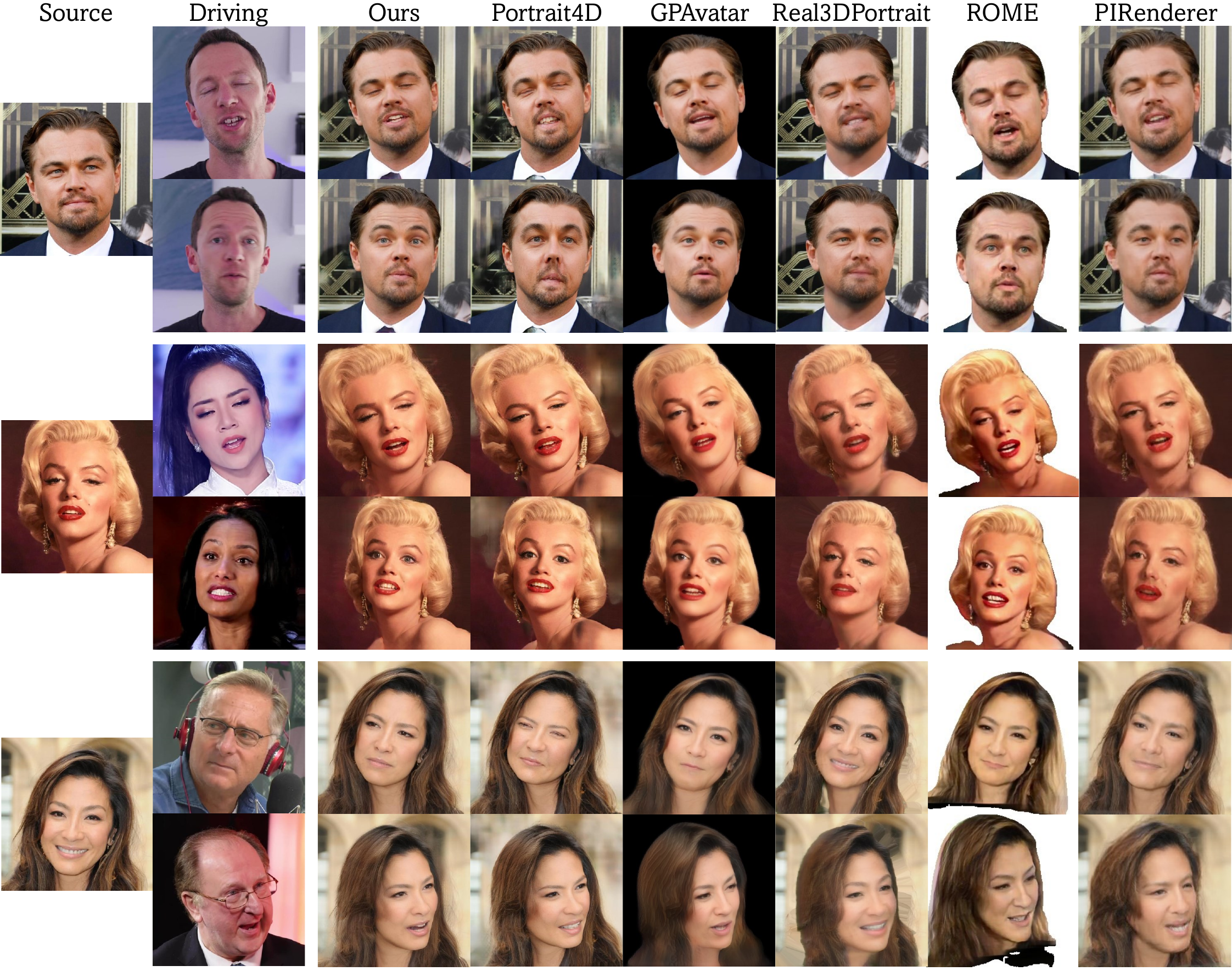}
	\caption{Qualitative comparison using in-the-wild sources and VFHQ drivings.}
	\label{fig:compare}
\end{figure*}

\subsection{Comparisons}\label{sec:compare}
\paragraph{\textbf{Baselines.}} We compare with existing one-shot video-based head reenactment methods, including 2D-based approaches: PIRenderer~\cite{ren2021pirenderer}, Face-vid2vid~\cite{wang2021one}, and StyleHEAT~\cite{yin2022styleheat}; and 3D-aware ones: ROME~\cite{khakhulin2022realistic}, OTAvatar~\cite{ma2023otavatar}, HideNeRF~\cite{li2023one}, Real3DPortrait~\cite{ye2024real3d}, GPAvatar~\cite{chu2024gpavatar}, GOHA~\cite{li2023generalizable}, and Portrait4D~\cite{deng2023portrait}.

\paragraph{\textbf{Quantitative results.}} We first conduct self- and cross-ID reenactment on a test split of 100 video clips in VFHQ, each with around 200 frames. To measure image synthesis quality, we use LPIPS~\cite{zhang2018unreasonable} and Fr\'echet Inception Distances (FID)~\cite{heusel2017gans} between the synthesized results and the ground truth. For identity similarity, we calculate cosine distance of the face recognition features~\cite{deng2019arcface} between the synthesis results and the source appearance. For expression, we use Average Expression Distance (AED)~\cite{lin20223d} measured by a 3DMM estimator~\cite{deng2019accurate}. We also 
compute Average Pose Distance (APD, ×1000 by default)~\cite{chan2021efficient} to measure the pose control accuracy.
As shown in Tab.~\ref{tab:main_self}, our method largely outperforms previous methods in terms of reconstruction fidelity (\ie, LPIPS and FID), identity similarity (\ie, ID), and expression and pose control accuracy (\ie, AED and APD).

Our cross-ID AED is slightly higher than a few approaches. We argue that this commonly used metric, reliant on a 3DMM reconstructor~\cite{deng2019accurate}, may not adequately capture the similarity of expressions across identities with distinct shapes. To better evaluate cross-ID expression similarity, we report an AED$_f$ metric based on the feature of an emotion classifier~\cite{danvevcek2022emoca} in Tab.~\ref{tab:cross-exp}. Moreover, to fairly evaluate the cross-ID lip motion accuracy, we utilize driving sequences from CelebV-Text~\cite{yu2023celebv} and measure the SyncNet~\cite{chung2017out} score between the reenacted results and the corresponding audios of the driving frames. As shown in Tab.~\ref{tab:cross-exp}, our method achieves the best expression and lip motion accuracy, compared to other methods with good cross-ID performance (indicated by Tab.~\ref{tab:main_self}). 

\begin{figure*}[t]
	\small
	\centering
	\includegraphics[width=1.0\textwidth]{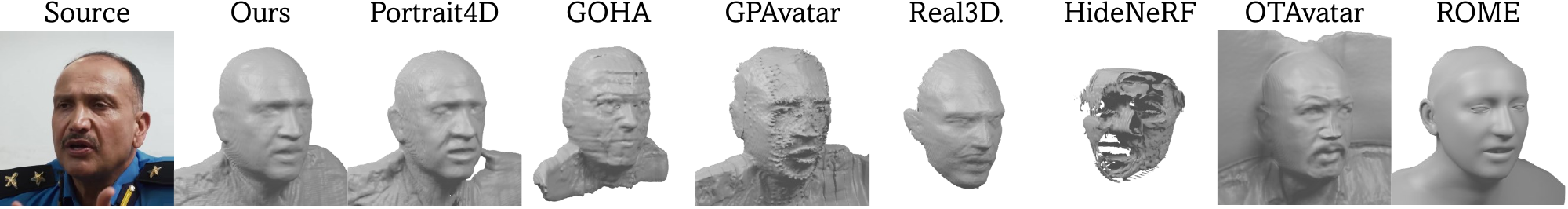}
	\caption{Reconstructed head geometries using different 3D-aware methods.}
	\label{fig:compare_geo}
\end{figure*}

\paragraph{\textbf{Qualitative results.}} Figure~\ref{fig:compare} shows visual comparisons between different methods. Compared to the alternatives, our method excels at reconstructing the appearance details of the source images while capturing subtle facial movements of mouth, eyes, eyebrows, etc., in the driving frames. Besides, our method can well maintain the identity consistency under large pose variations. 

Figure~\ref{fig:compare_geo} visualizes the reconstructed head geometries (via Marching Cubes) of a given source image using different 3D-aware methods. Our method best preserves the overall head structure, which is the key for a good 3D consistency. Note that for HideNeRF~\cite{li2023one}, we failed to obtain a reasonable geometery. 

\paragraph{\textbf{User study.}} For a more comprehensive evaluation, we ask $22$ participants to select the best method in terms of image synthesis quality and expression similarity while inspecting 20 cross-ID reenactment results synthesized by different methods, respectively. As shown in Tab.~\ref{tab:user-study}, our method demonstrates significant advantages compared to the others. Raw results are in Sec.~\ref{sec:user_study_raw}.

\subsection{Ablation Study}\label{sec:ablation}

\begin{figure*}[t]
	\small
	\centering
	\includegraphics[width=1.0\textwidth]{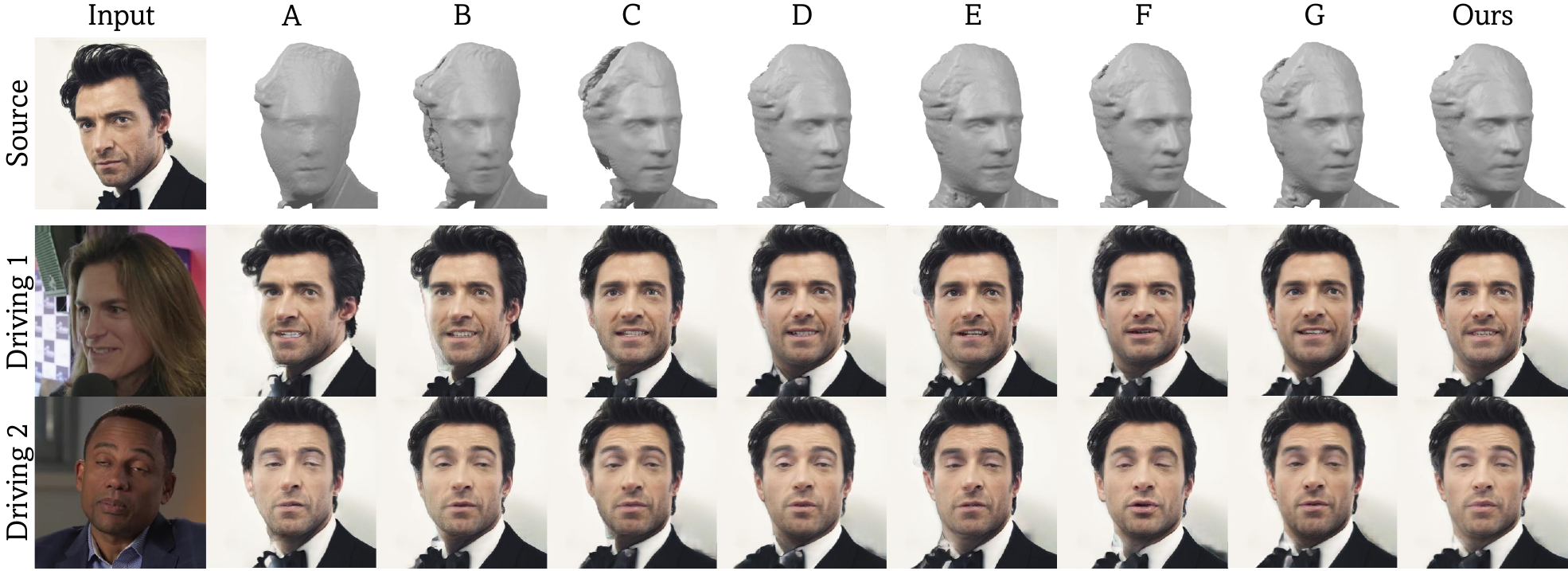}
	\caption{Ablation study on different configurations. The first row visualizes the reconstructed geometries of the source. The bottom two rows depict cross-ID reenactment results rendered at a frontal view given two different drivings.}
	\label{fig:ablation}
\end{figure*}

We conduct ablation studies to validate our proposed strategies. Table~\ref{tab:ablation} reports the quantitative results following the experiment setting in Sec.~\ref{sec:compare}. We introduce an extra metric (\ie, Cons.) to evaluate identity consistency of the synthesized results given different drivings. Specifically, for each source, we use $50$ drivings with different identities and poses for reenactment and compute the average cosine similarity between identity features~\cite{deng2019arcface} extracted from the synthesized results at frontal view. A higher Cons. indicates better disentanglement of motion with other information (\eg, pose and ID) in the drivings. In Fig.~\ref{fig:ablation}, we visualize the reconstructed source geometries and \emph{frontal-view} reenacted results under two different drivings. All configurations are trained to see $5$M images by default.

\begin{table*}[t]
\centering
\caption{Ablation study of our method conducted on the test split of VFHQ.}
% \resizebox{\textwidth}{20mm}{
\begin{adjustbox}{width=0.97\columnwidth,center}
\begin{tabular}{l|cccccc|cccc|ccccc}
\toprule

\multicolumn{1}{c}{} &\multicolumn{6}{c}{\multirow{1}{*}{Configuration}} & \multicolumn{4}{c}{Self reenactment}  & \multicolumn{5}{c}{Cross-ID reenactment} \\ 
\cline{2-16} 
 \multicolumn{1}{c}{} & Init. & $I_d(\bm \theta_p)$  &  $I_d(\bm \theta_q)$ & Mot. & 3D Syn. & Data& LPIPS $\downarrow$  & ID $\uparrow$ & AED $\downarrow$ & APD $\downarrow$ & ID $\uparrow$ & AED $\downarrow$ & AED$_f$ $\downarrow$ & APD $\downarrow$ & Cons. $\uparrow$ \\
 \midrule
 A & Rand. & & & $E_{mot}$ & $\mathrm{\Psi}_{3d}$ & 10K & 0.216  & 0.874 & 0.021  & 0.752 & 0.776 & 0.167 & 0.213 & 1.887 & 0.822 \\ 
 B & $\mathrm{\Psi}_{3d}$ & & & $E_{mot}$ & $\mathrm{\Psi}_{3d}$ & 10K & 0.213 & 0.879 & 0.018  & 0.755 & 0.718 & 0.161 & 0.185 &  0.796 & 0.813 \\ 
  C & $\mathrm{\Psi}_{3d}$ & \checkmark & & $E_{mot}$ & $\mathrm{\Psi}_{3d}$ & 10K &  0.223 & 0.873 & 0.020 & 1.251 & 0.757 & 0.165 & 0.190 & 1.436 & 0.913 \\ 
  D & $\mathrm{\Psi}_{3d}$ &  & \checkmark & $E_{mot}$ & $\mathrm{\Psi}_{3d}$ & 10K &  0.219 & 0.875 & 0.019  & 0.739 & 0.712 & 0.156 & 0.185 & 0.688 & 0.805\\ 
  \midrule
  E & Rand. & \checkmark & \checkmark & $E_{mot}$ & $\mathrm{\Psi}_{3d}$ & 10K &  0.232 & 0.865 &0.023 & 0.967 & 0.777 & 0.172 & 0.207 & 1.257 & 0.929\\ 
  F & $\mathrm{\Psi}_{3d}$ & \checkmark & \checkmark & 3DMM & $\mathrm{\Psi}_{3d}$ & 10K & 0.221  & 0.873 & 0.018 & 1.074 & 0.704 & 0.139 & 0.202 & 0.993 & 0.811 \\ 
  G & $\mathrm{\Psi}_{3d}$ & \checkmark & \checkmark & $E_{mot}$ & \cite{deng2023portrait} & 10K & 0.225  & 0.867 &  0.021 & 1.476 & 0.744 & 0.163 & 0.188 & 1.067 & 0.912\\ 
  \midrule
  H & $\mathrm{\Psi}_{3d}$ & \checkmark & \checkmark & $E_{mot}$ & $\mathrm{\Psi}_{3d}$ & 1K & 0.273  & 0.580 & 0.050  & 0.603 & 0.470 & 0.141 & 0.181 & 0.700 & 0.864\\ 
  I & $\mathrm{\Psi}_{3d}$ & \checkmark & \checkmark & $E_{mot}$ & $\mathrm{\Psi}_{3d}$ & 2K &  0.257 & 0.707 & 0.038  & 0.586 & 0.567 & 0.144 & 0.178 & 0.810 & 0.879 \\ 
 J & $\mathrm{\Psi}_{3d}$ & \checkmark & \checkmark & $E_{mot}$ & $\mathrm{\Psi}_{3d}$ & 5K &  0.236 & 0.818 & 0.026  & 1.527 & 0.678 & 0.154 & 0.187 & 1.039 & 0.904\\ 
 \midrule
 K(Ours) & $\mathrm{\Psi}_{3d}$ & \checkmark & \checkmark & $E_{mot}$ & $\mathrm{\Psi}_{3d}$ & 10K & 0.224  & 0.873 & 0.019  & 0.758 & 0.743 & 0.162 & 0.186 & 0.943 & 0.912\\ 
  L(Ours*) & $\mathrm{\Psi}_{3d}$ & \checkmark & \checkmark & $E_{mot}$ & $\mathrm{\Psi}_{3d}$ & 10K & 0.224  & 0.873 & 0.019  & 0.580 & 0.736 & 0.161 & 0.185 & 0.858 & 0.908\\ 
\bottomrule
\end{tabular}
\end{adjustbox}
% }

\label{tab:ablation}
\end{table*}

\paragraph{\textbf{Initialization.}} We first validate the efficacy of weight initialization using the pre-trained $\mathrm{\Psi}_{3d}$ (\ie, A \emph{vs.} B and E \emph{vs.} Ours). As shown in Tab.~\ref{tab:ablation}, using $\mathrm{\Psi}_{3d}$ as initialization yields better self-reenactment performance as well as more accurate expression and pose control in the cross-ID setting. Moreover, Fig.~\ref{fig:ablation} depicts that $\mathrm{\Psi}_{3d}$ helps with more reasonable geometry learning. 

\paragraph{\textbf{Pseudo multi-view data.}} We evaluate the influence of the pseudo multi-view videos by removing $I_d(\bm \theta_p)$ or $I_d(\bm \theta_q)$ during training (B \emph{vs.} C \emph{vs.} D \emph{vs.} Ours). Leveraging $I_d(\bm \theta_p)$ for extracting the motion embedding largely improves the identity consistency (\ie Cons.) with only a slight drop of the overall reenactment performance. This validates that $I_d(\bm \theta_p)$ can help eliminate the non-expression information (\eg, pose) within the pre-trained 2D motion embedding~\cite{wang2023progressive}, which is also in accordance with the higher ID metric in the cross-ID setting. Using $I_d(\bm \theta_q)$ as multi-view supervision facilitates geometry learning and reconstruction of unseen regions, as indicated by the lower APD metrics and shown in Fig.~\ref{fig:ablation}. Utilizing both of them yields our method with faithful 3D reconstruction ability and competitive reenactment performance. More analysis are in Sec.~\ref{sec:analysis}.
 
\paragraph{\textbf{Motion representation.}} We further replace our motion embedding from~\cite{wang2023progressive} with motion-related 3DMM parameters extracted following Sec.~\ref{sec:3d} (\ie, F \emph{vs.} Ours). As shown in Tab.~\ref{tab:ablation}, using 3DMM parameters for motion control is detrimental to identity preservation in the cross-ID setting, due to a known identity leakage issue of linear 3DMMs~\cite{jiang2019disentangled,deng2023portrait}. While it has a lower AED metric (which is also a 3DMM-based metric without considering identity entanglement), it performs worse in terms of AED$_f$ and cannot faithfully capture detailed facial expressions as shown in Fig.~\ref{fig:ablation}. Notably, while leveraging the pre-trained 2D motion embedding can yield better cross-ID reenactment result, it is non-trivial to obtain reasonable 3D head geometries without our pseudo multi-view data for training (as indicated by config.~B). See Sec.~\ref{sec:analysis} for more visual analysis.

\paragraph{\textbf{3D Synthesizer.}} We also compare with a baseline which adopts portrait4D~\cite{deng2023portrait} for synthesizing multi-view training data (G \emph{vs.} Ours). Our 3D head synthesizer $\mathrm{\Psi}_{3d}$ facilitates expression control and geometry learning as indicated by the lower AED, AED$_f$, and APD in Tab.~\ref{tab:ablation}, respectively, due to its better structure preservation ability during novel-view synthesis (see Fig.~\ref{fig:static_compare}). 

\paragraph{\textbf{Number of training data.}} We evaluate our method using different numbers of training videos (H \emph{vs.} I \emph{vs.} J \emph{vs.} Ours). Table~\ref{tab:ablation} shows that increasing the number of training data largely improves the reconstruction fidelity and identity preservation ability. We believe a even higher performance is reachable by further scaling up the training data. However, we observe that the models learned with fewer data have lower AED and APD in the cross-ID setting. We speculate that this may be due to a more severe information-leakage issue of the motion embedding when trained with limited data, which misleads the corresponding metrics (as indicated by the lower Cons. metric).

\paragraph{\textbf{Hyper parameters.}} We slightly tune the hyper parameters for a better performance (Ours \emph{vs.} Ours*). By default, all parameters within the model share the same learning rate of $1e-4$. We scale up that of the motion-related layers by a factor of $2.5$, and train the model to see $6$M images, which empirically lowers the APD metric with minor influence to the remaining, yielding our final model. 

\section{Conclusion}
In this paper, we presented a novel learning approach for synthesizing lifelike 4D head avatars from a single appearance image. The key intuition is to turn existing monocular videos into pseudo multi-view ones to enable learning an image-to-4D-head synthesizer in a data-driven manner instead of leveraging 3DMM-based inductive bias with limited expressiveness. 
% To do so, we first learn a 3D head synthesizer using multi-view synthetic images, and then leverage it to synthesize novel views of the monocular videos for cross-view self-reenactment learning of the 4D head synthesizer. 
Extensive experiments have demonstrated the superiority of our method over the prior arts in terms of reconstruction fidelity, geometry consistency, and reenactment accuracy. We hope our method will inspire future works towards exploring a better incorporation of 3D priors with 2D data for more realistic head avatar synthesis.

% \paragraph{\textbf{Limitations and future works.}} Our model may produce fewer details for unseen regions due to its deterministic property. Besides, it cannot faithfully mimic extreme expressions restricted by the capability of the pre-trained motion embedding. We refer to Sec.~\ref{sec:limitation_more} for a thorough discussion and possible improvements in the future.

% \paragraph{\textbf{Ethics consideration.}} The proposed method is intended for virtual communication and entertainment. However, a misuse of it for deceptive content creation can be harmful and should be strictly prohibited. Currently, the synthesized results contain certain visual artifacts which could help with deepfake detection. 

% \clearpage  % TODO REVIEW/FINAL: This \clearpage needs to be removed from both review and camera-ready versions.

\bibliographystyle{splncs04}
\bibliography{main}

\clearpage
\appendix

\renewcommand{\thesection}{\Alph{section}}
\renewcommand{\thefigure}{\Roman{figure}}
\renewcommand{\thetable}{\Roman{table}}
\renewcommand{\theequation}{\Roman{equation}}
\setcounter{section}{0}
\setcounter{figure}{0}
\setcounter{equation}{0}

\section{More Implementation Details}\label{sec:implement}

\paragraph{\textbf{Learning the 3D synthesizer $\mathrm{\Psi}_{3d}$.}} We use GenHead~\cite{deng2023portrait} trained on FFHQ~\cite{karras2019style} at $512^2$ to synthesize multi-view supervisions for learning $\mathrm{\Psi}_{3d}$. For high-quality data synthesis, we extract FLAME parameters from real images in FFHQ and VFHQ~\cite{xie2022vfhq} as input conditions to GenHead. For camera pose, we follow~\cite{deng2023portrait} to sample it from a pre-defined uniform distribution,
with pitch from $[-0.25,0.65]$ $rad$, yaw from $[-0.78,0.78]$ $rad$, and roll from $[-0.25,0.25]$ $rad$. We also sample the camera radius from $[3.65,4.45]$, and the camera look-at position from $[-0.01,0.01]\times[-0.01,0.01]\times[0.02,0.04]$. The camera intrinsics are fixed with a $12^\circ$ field of view (fov). 

We use the balancing weights of~\cite{deng2023portrait} for each loss term in Eq.~\eqref{eq:training_3d} during training. At the first $1$M images, we deactivate the adversarial loss $\mathcal{L}_{adv}$ and fix the learnable network parameters within the renderer $\mathcal{R}$, where we use an MLP for radiance decoding and a StyleGAN2-based~\cite{karras2020analyzing,chan2021efficient} super-resolution module for synthesizing the final images of high resolution. After seeing $1$M images, we activate $\mathcal{L}_{adv}$ and update all parameters in $\mathcal{R}$ along with $\mathrm{\Psi}_{3d}$. $\mathcal{L}_{depth}$ and $\mathcal{L}_{T}$ are removed at this stage. Besides, we use a neural rendering resolution of $64^2$ at the first $1$M images, and linearly increase it to $128^2$ during the next $1$M-image period, while the final image resolution is consistently set to $512^2$. We sample $48$ coarse points and $48$ fine points per ray following~\cite{chan2021efficient} for volume rendering. We adopt an Adam optimizer~\cite{kingma2015adam} with a learning rate of $1e-4$ and a batch size of $32$, and train $\mathrm{\Psi}_{3d}$ to see $10$M images in total, which takes 8 days on 8 Tesla A100 GPUs with 80GB memory.

\paragraph{\textbf{Learning the 4D synthesizer $\mathrm{\Psi}$.}} We initialize $\mathrm{\Psi}$ using the pre-trained weights of $\mathrm{\Psi}_{3d}$ except the motion-related cross-attention layers that are initialized at random. We then use a training split of $10$K video clips from VFHQ~\cite{xie2022vfhq} and sample $50$ frames per clip to learn $\mathrm{\Psi}$. During each iteration,  we randomly choose $2$ frames in a video clip as the source and driving, respectively, and send the driving image to the pre-trained $\mathrm{\Psi}_{3d}$ to obtain its pseudo multi-view frames, where we use a similar camera distribution for rendering as described in the above section. We adopt the balancing weights used in learning $\mathrm{\Psi}_{3d}$ for the four loss terms in Eq.~\eqref{eq:training_4d}. We use a neural rendering resolution of $128^2$ throughout the training process. We train $\mathrm{\Psi}$ to see $6$M images, using Adam with a learning rate of $2.5e-4$ for the motion-related layers and $1e-4$ for the remaining, and a batch size of $32$, which also takes 8 days on 8 A100 GPUs. 

\paragraph{\textbf{Neck pose rotation.}} We do not utilize the motion embedding~\cite{wang2023progressive} to model the neck pose. Instead, we introduce an explicit 3D rotation field to tackle it following~\cite{deng2023portrait}. More specifically, we pre-define a voxel grid of $24^3$ and compute the 3D rotation of each grid point via a Surface Field (SF)~\cite{bergman2022generative} derived from a FLAME mesh at run time. Then, for each 3D point, we obtain its rotation via interpolating those of its nearest grid points. This way, the 3D rotation for arbitrary number of points can be efficiently calculated. Besides, the control of facial expressions and neck pose can be well disentangled.

\begin{figure}[t]
	\small
	\centering
	\includegraphics[width=0.98\columnwidth]{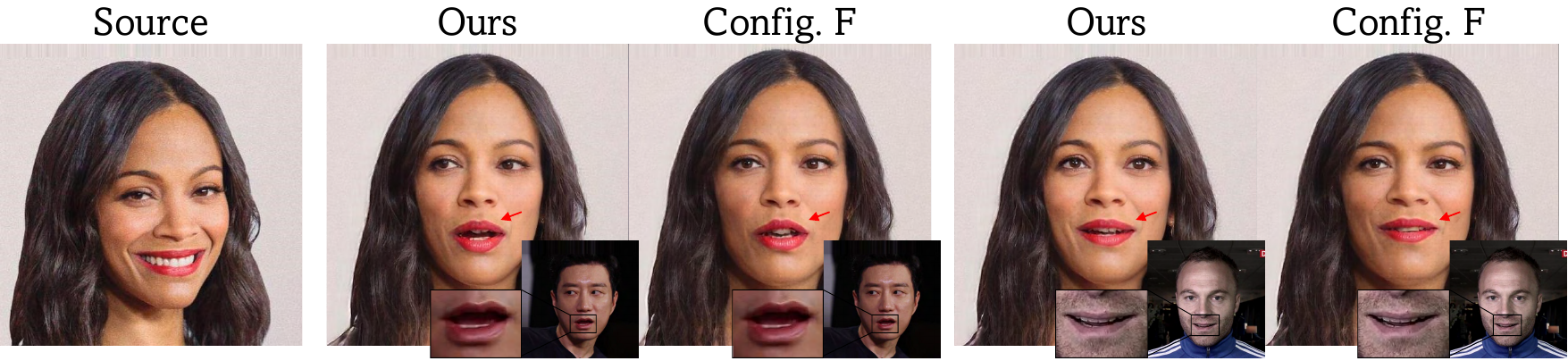}
    \caption{Comparisons between configurations using different motion representations. Ours using implicit 2D motion embedding learned from large-scale video data surpasses config.~F using 3DMM at nuanced expression control (\eg, see lip and teeth motions).}
    \label{fig:embedding}
\end{figure}

\section{More Synthesis Results}\label{sec:results_more}
Please see the \emph{project page} for more head synthesis results of our method, where we use in-the-wild images collected from the Internet as the source and video sequences from CelebV-Text~\cite{yu2023celebv} as the driving.

\section{More Comparisons}\label{sec:compare_more}
We provide more comparisons between our method and the prior art in the \emph{project page}. We use images from VFHQ test set as the source and those from CelebV-Text as the driving. 

\section{More Visual Analysis}\label{sec:analysis}
In this section, we offer a more comprehensive analysis of the advantages of our proposed framework, which combines implicit 2D motion representation with multi-view training data. 

As mentioned in the main paper, we employ an implicit expression representation learned from 2D data (\ie, PD-FGC~\cite{wang2023progressive}) to capture subtle motions difficult to be described using existing 3DMMs. As illustrated in Fig.~\ref{fig:embedding}, our data-driven motion embedding excels at modeling the nuanced mouth movements that occur during speech. In contrast, 3DMMs struggle to accurately represent the relative positions between lips and teeth.

\begin{figure}[t]
	\small
	\centering
	\includegraphics[width=0.98\columnwidth]{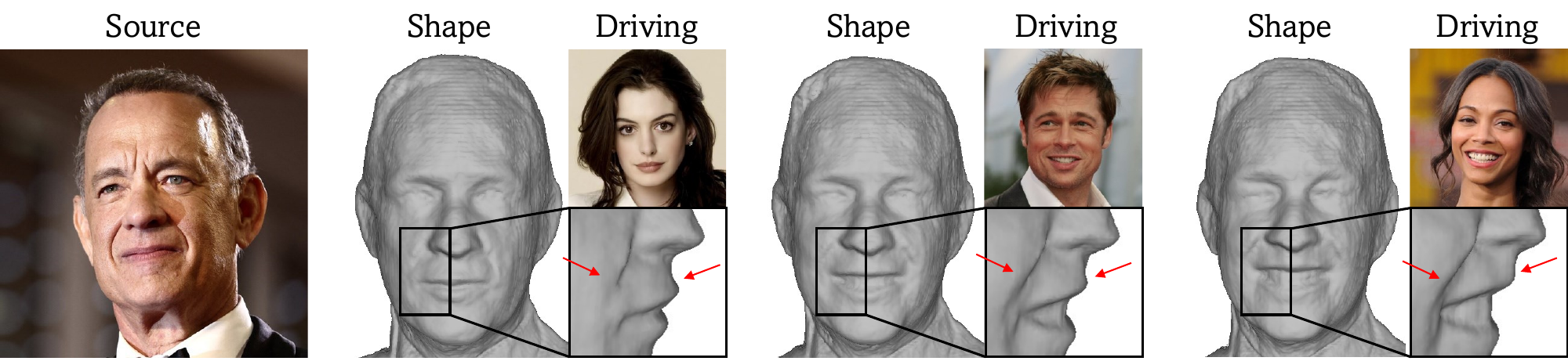}
    \caption{Our 2D motion embedding, even though it is learned from soley 2D videos, effectively captures reasonable 3D geometry changes.}
    \label{fig:3d}
\end{figure}

Additionally, even though PD-FGC is a 2D motion embedding, it well records 3D geometry information. Figure~\ref{fig:3d} shows that our motion embedding effectively captures subtle 3D geometry changes around cheeks and lips given different 2D driving images. We contend that this is because most 3D geometry clues can be inferred from 2D appearances. 

\begin{figure}[t]
	\small
	\centering
	\includegraphics[width=0.96\columnwidth]{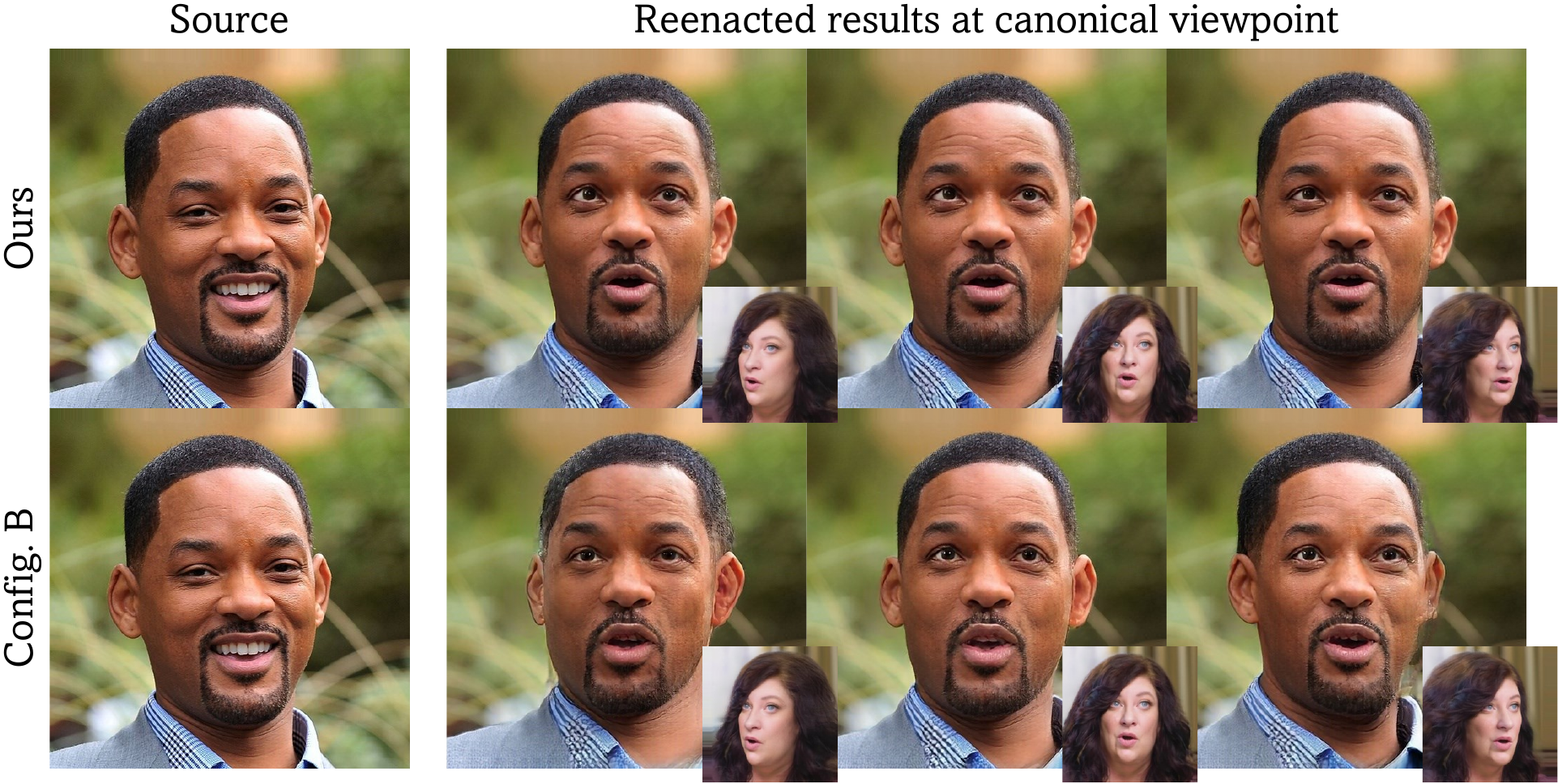}
    \caption{By utilizing multi-view training data, we effectively remove pose-related information from our 2D motion embedding, achieving enhanced 3D consistency while preserving the original capability to represent detailed motion.}
    \label{fig:consistency}
\end{figure}

Furthermore, although the initial PD-FGC representation includes pose-related information due to its 2D nature, our learning approach using multi-view training data enforces 3D consistency while preserving detailed expression information. Figure~\ref{fig:consistency} demonstrates that our method with multi-view supervision retains the ability to reconstruct subtle expressions compared to config. B using monocular video supervision, while also providing improved 3D consistency.

\section{User Study}\label{sec:user_study_raw}
\begin{figure*}[p]
	\small
	\centering
	\includegraphics[width=1.0\textwidth]{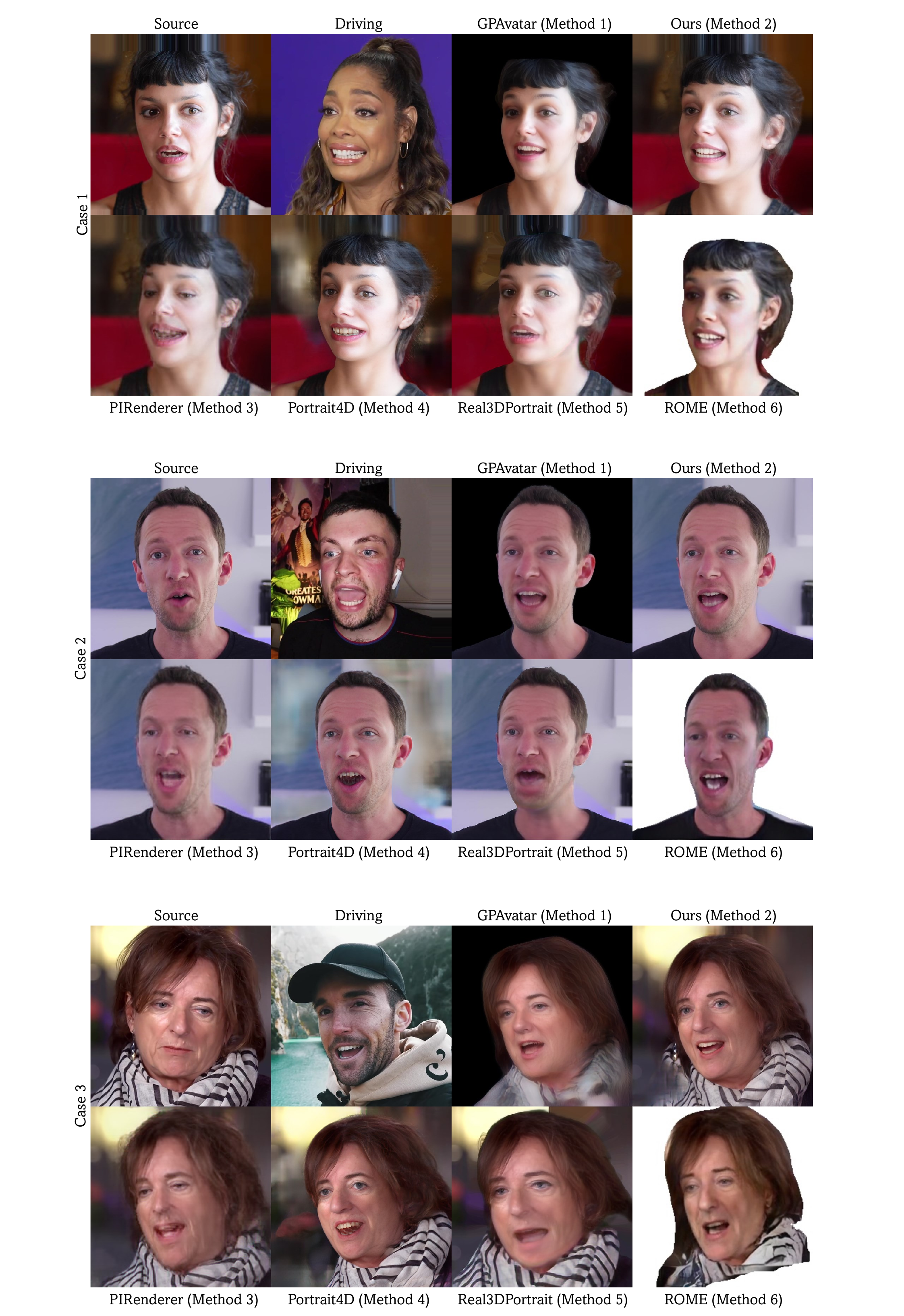}
	\caption{User study cases.}
	\label{fig:user1}
\end{figure*}

\begin{figure*}[p]
	\small
	\centering
	\includegraphics[width=1.0\textwidth]{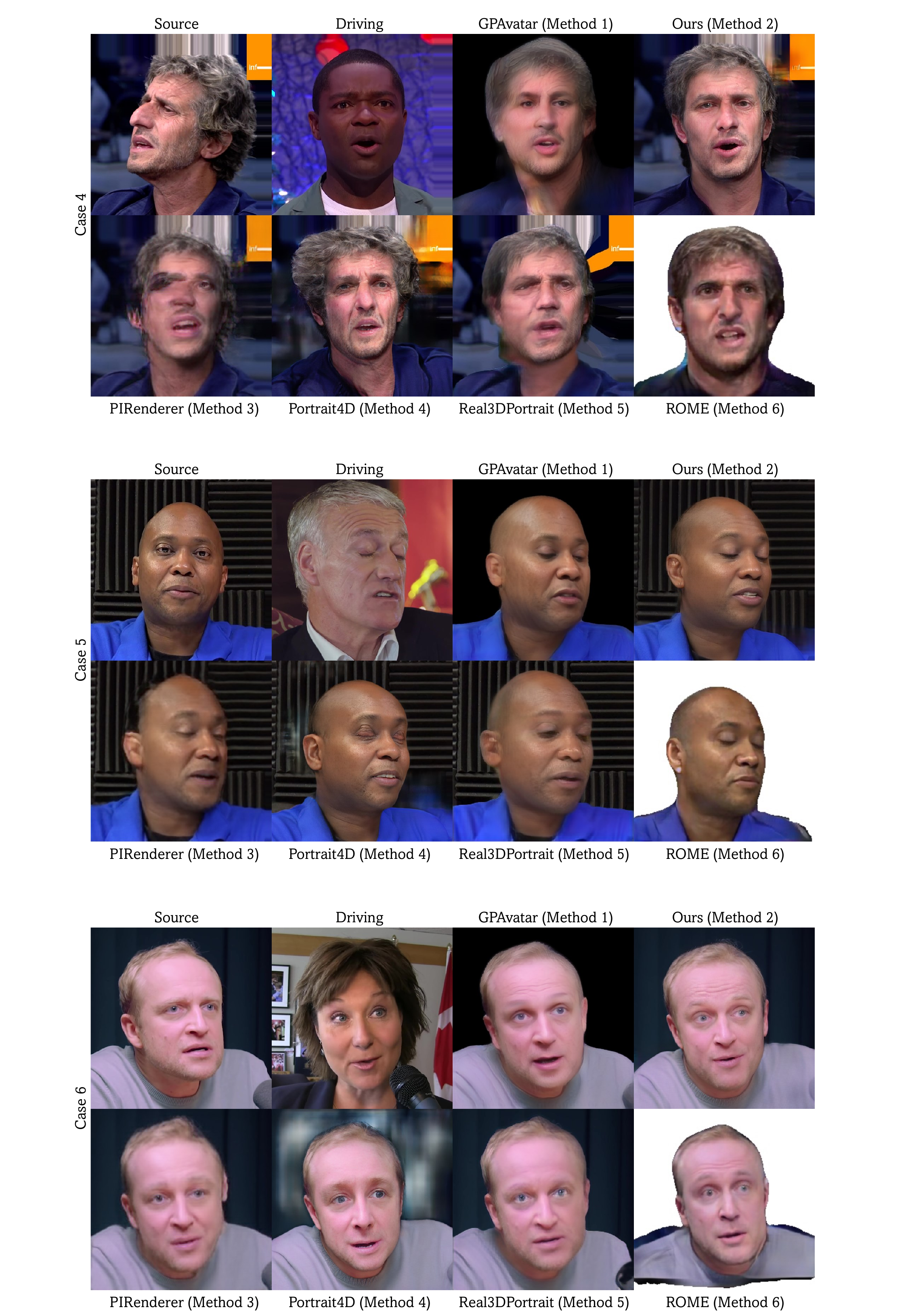}
	\caption{User study cases.}
	\label{fig:user2}
\end{figure*}

\begin{figure*}[p]
	\small
	\centering
	\includegraphics[width=1.0\textwidth]{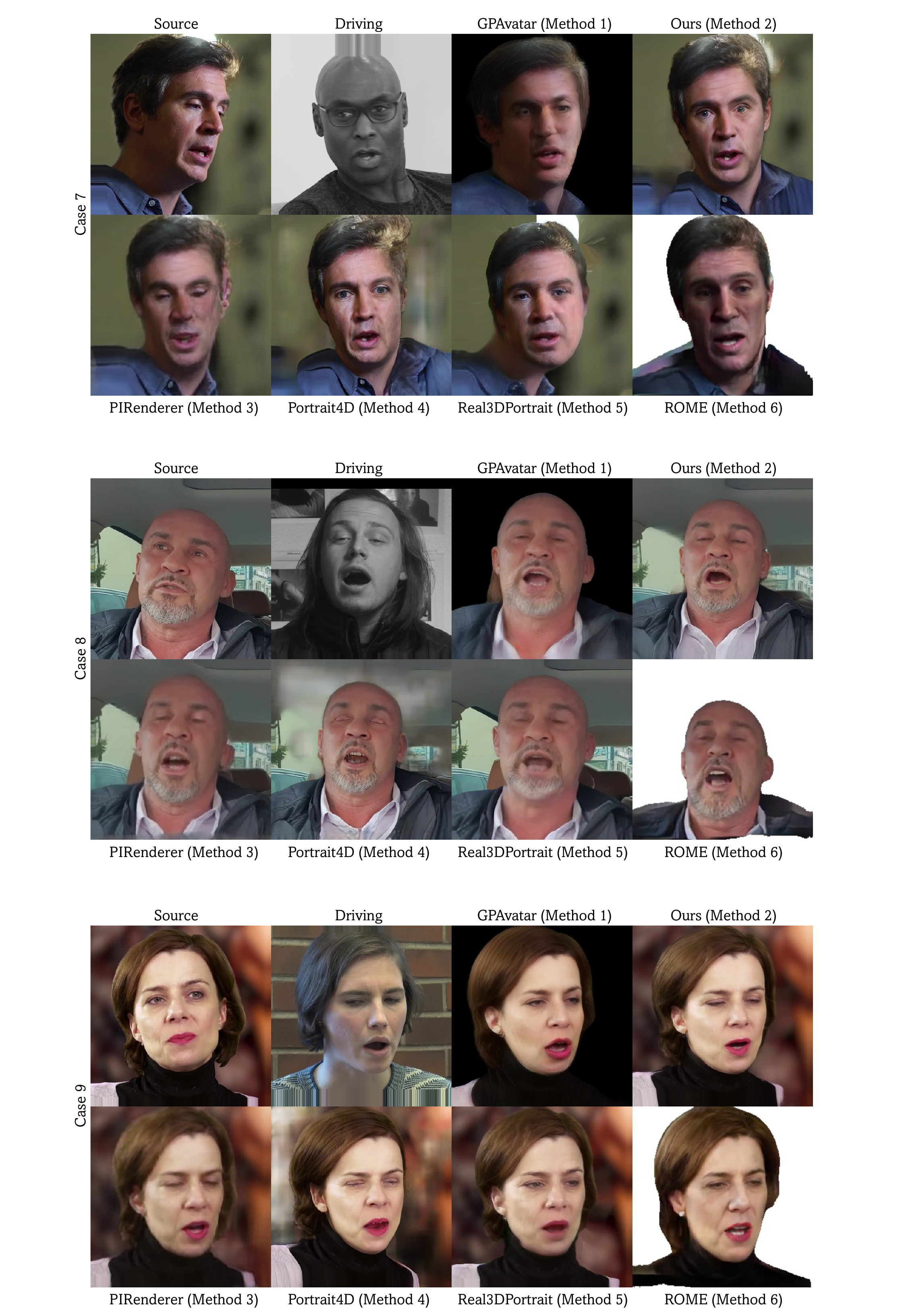}
	\caption{User study cases.}
	\label{fig:user3}
\end{figure*}

\begin{figure*}[p]
	\small
	\centering
	\includegraphics[width=1.0\textwidth]{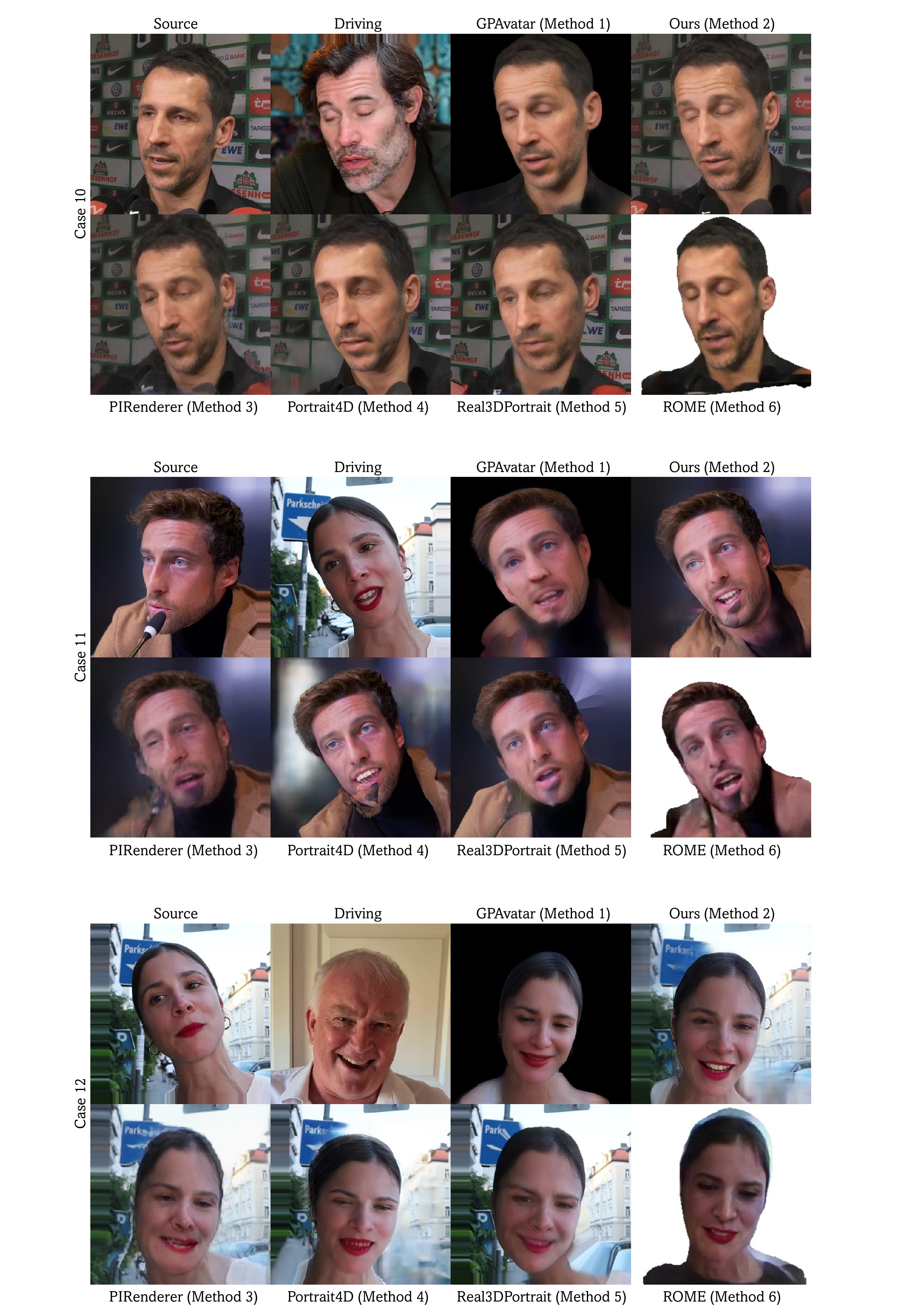}
	\caption{User study cases.}
	\label{fig:user4}
\end{figure*}

\begin{figure*}[p]
	\small
	\centering
	\includegraphics[width=1.0\textwidth]{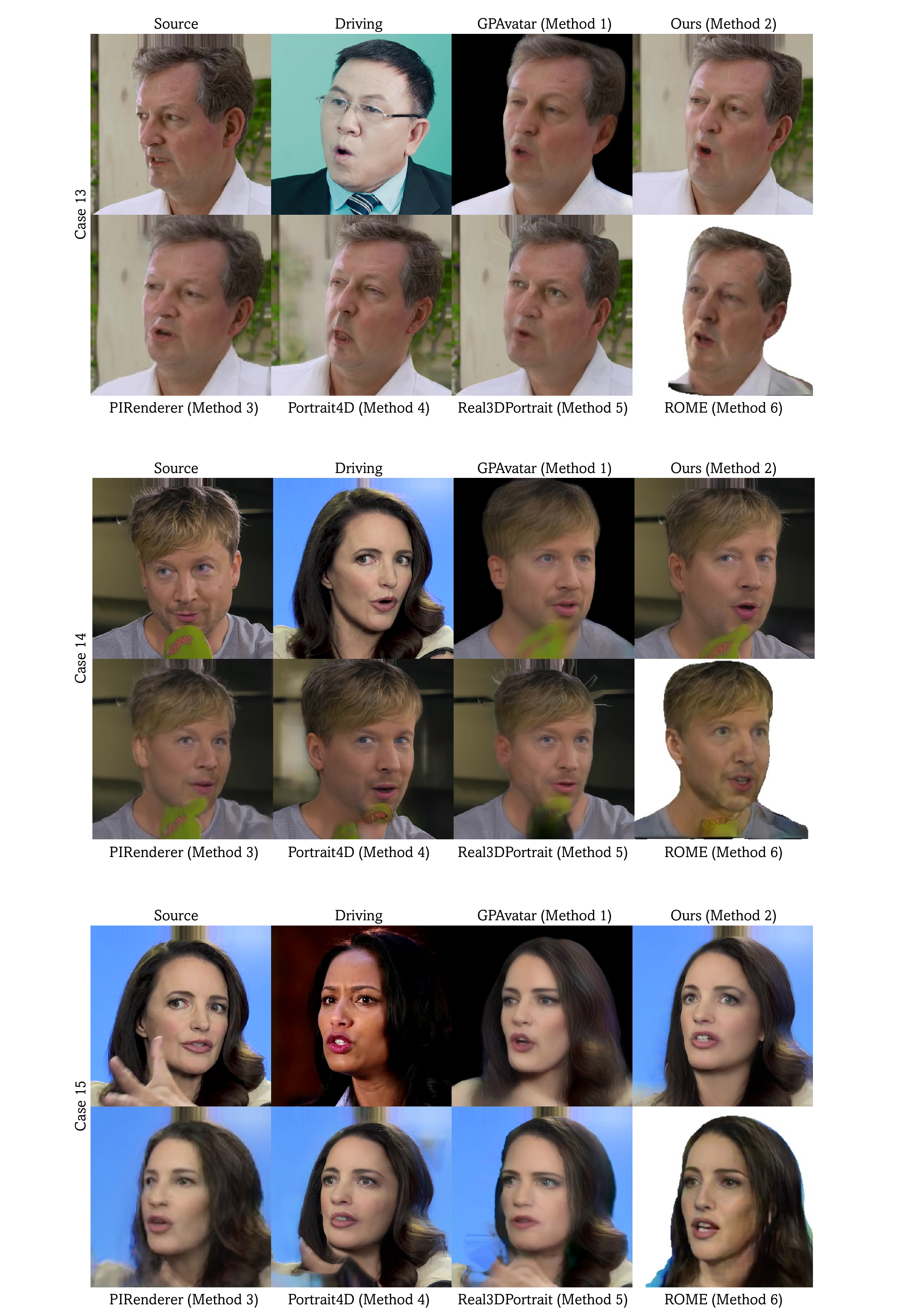}
	\caption{User study cases.}
	\label{fig:user5}
\end{figure*}

\begin{figure*}[p]
	\small
	\centering
	\includegraphics[width=1.0\textwidth]{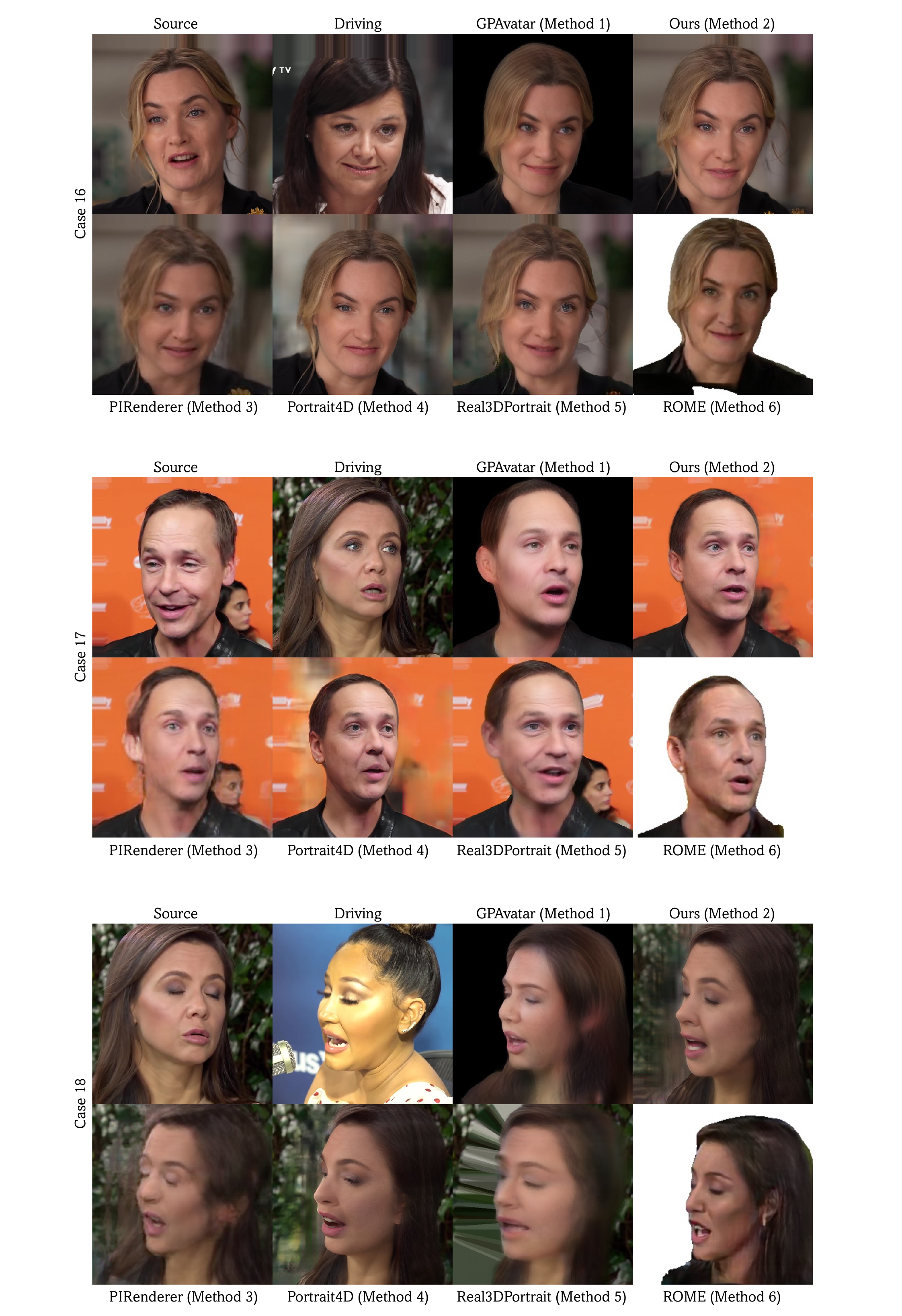}
	\caption{User study cases.}
	\label{fig:user6}
\end{figure*}

\begin{figure*}[p]
	\small
	\centering
	\includegraphics[width=1.0\textwidth]{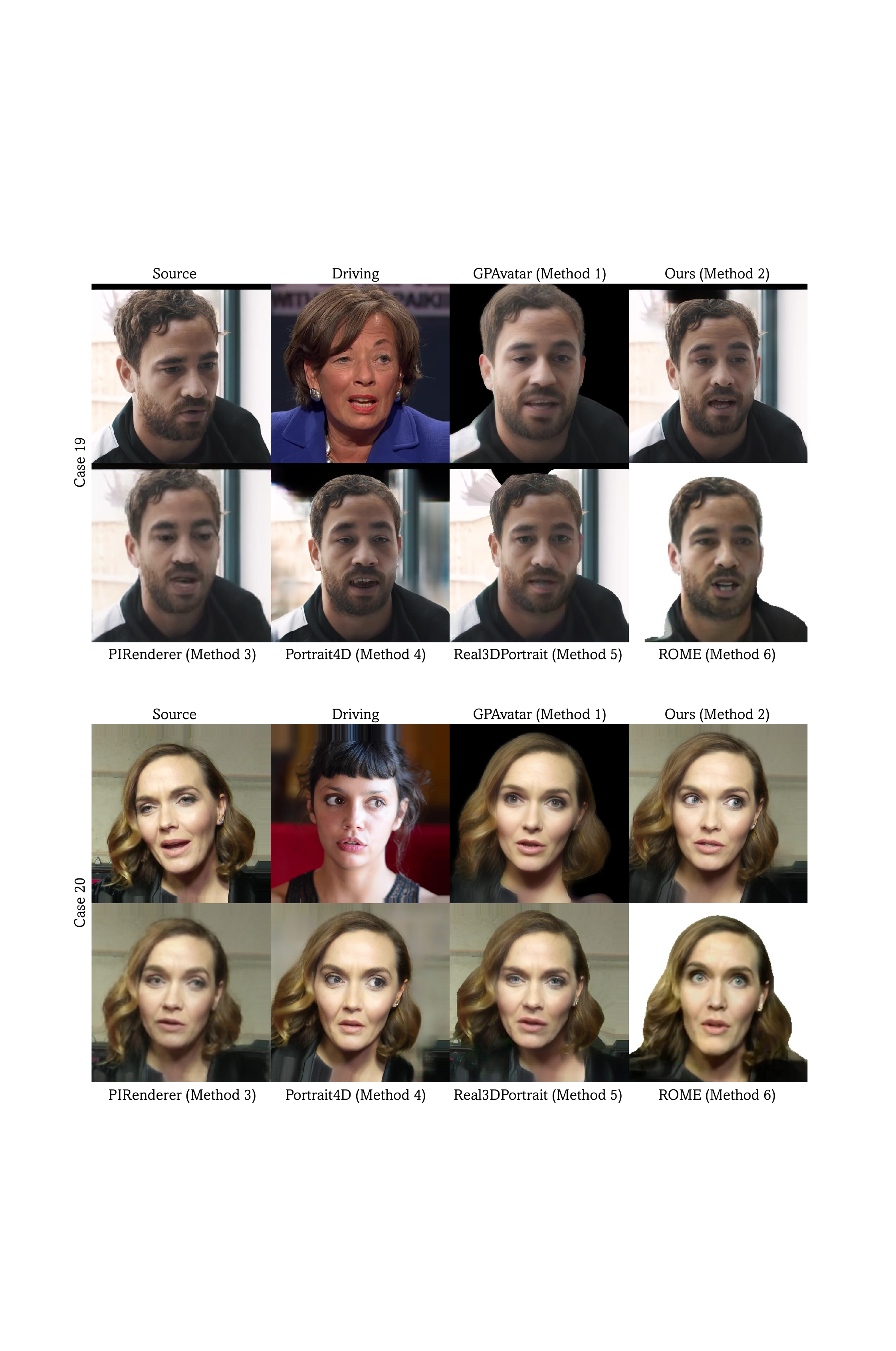}
 \vspace{-85pt}
	\caption{User study cases.}
	\label{fig:user7}
\end{figure*}

We provide the $20$ test cases used for the user study in Fig.~\ref{fig:user1} to~\ref{fig:user7}  and the original voting results in Tab.~\ref{tab:user1} and~\ref{tab:user2}.

\section{Limitations and Future Works}\label{sec:limitation_more}
Despite the high-quality synthesis results compared to previous approaches, our method still has several limitations. 

First, we observe that when rendering the synthesized results at novel views, the unseen regions in the original source image usually have fewer details compared to the visible regions. We conjecture that this is due to the deterministic property of our 4D head synthesizer, where it tends to predict an ``average'' expectation of all possible results. We believe integrating the current learning framework with a stochastic generative model (\eg, a Diffusion Model~\cite{rombach2022high}) can alleviate this issue. 

Second, our current motion embedding from~\cite{wang2023progressive} is learned using video data from Voxceleb2~\cite{Chung18b} which may lack extreme expressions. As a result, our method may not well capture exaggerated facial movements such as puffing out the cheeks or poking out the tongue. However, since our learning framework does not require an end-to-end training with the motion encoder, it is possible to re-train the motion embedding using more expressive data or replace it with more advanced motion embedding for better expression imitation in the future. 

Finally, as a learning-based approach, our method may produce inferior results for out-of-distribution data and can have slightly different performances for identities of different races. We plan to collect training data of higher diversity and coverage to alleviate this problem. Besides, extending our approach to handle upper-body or full-body avatar synthesis is also an interesting future direction to be explored.

\section{Ethics Consideration}\label{sec:ethics}
The proposed method is intended for virtual communication and entertainment. However, a misuse of it for deceptive content creation can be harmful and should be strictly prohibited. Currently, the synthesized results contain certain visual artifacts which could help with deepfake detection. 

\clearpage


%%%%%%%%%%%%%%%%%%%%%%%%%%%%%%%%%%%%%%%%%%%%%%%%%%%%%%%%%%%%%%
\begin{table}[p] 
\caption{User study results on expression similarity. Each row shows the choice of one participant. \label{tab:user1}}
\begin{adjustbox}{width=1.0\columnwidth,center}
\begin{tabular} {cccccccccccccccccccc}
\toprule
case1 & case2 & case3 & case4 & case5 & case6 & case7 & case8 & case9 & case10 & case11 & case12 & case13 & case14 & case15 & case16 & case17 & case18 & case19 & case20 \\
\midrule
2     & 2     & 2     & 2     & 2     & 2     & 2     & 2     & 2     & 2      & 2      & 2      & 2      & 2      & 2      & 2      & 2      & 2      & 2      & 2      \\
2     & 1     & 2     & 2     & 1     & 2     & 2     & 2     & 1     & 1      & 2      & 2      & 1      & 2      & 2      & 1      & 2      & 2      & 2      & 2      \\
2     & 2     & 2     & 2     & 1     & 2     & 2     & 2     & 1     & 2      & 4      & 2      & 2      & 4      & 2      & 1      & 2      & 2      & 2      & 2      \\
4     & 6     & 2     & 1     & 2     & 2     & 2     & 2     & 6     & 1      & 2      & 2      & 2      & 4      & 5      & 3      & 2      & 2      & 6      & 2      \\
2     & 2     & 2     & 2     & 1     & 2     & 2     & 2     & 1     & 1      & 2      & 2      & 2      & 2      & 1      & 1      & 2      & 2      & 1      & 2      \\
2     & 2     & 2     & 2     & 2     & 2     & 2     & 1     & 1     & 2      & 2      & 2      & 2      & 4      & 2      & 1      & 2      & 2      & 2      & 2      \\
2     & 6     & 2     & 4     & 1     & 2     & 2     & 2     & 5     & 2      & 2      & 2      & 2      & 4      & 2      & 1      & 2      & 2      & 1      & 2      \\
2     & 2     & 2     & 2     & 2     & 2     & 2     & 2     & 2     & 2      & 2      & 2      & 2      & 2      & 2      & 1      & 2      & 2      & 1      & 2      \\
2     & 6     & 2     & 6     & 6     & 2     & 4     & 1     & 1     & 1      & 6      & 2      & 2      & 4      & 1      & 3      & 1      & 2      & 1      & 2      \\
2     & 2     & 2     & 2     & 2     & 2     & 2     & 2     & 2     & 2      & 2      & 2      & 2      & 2      & 2      & 2      & 2      & 2      & 2      & 2      \\
2     & 2     & 2     & 2     & 2     & 2     & 2     & 2     & 1     & 2      & 2      & 2      & 2      & 2      & 2      & 1      & 2      & 2      & 2      & 2      \\
2     & 1     & 2     & 1     & 1     & 2     & 2     & 2     & 1     & 1      & 2      & 2      & 2      & 2      & 2      & 1      & 2      & 2      & 2      & 2      \\
2     & 2     & 2     & 4     & 1     & 2     & 2     & 2     & 2     & 2      & 2      & 1      & 2      & 2      & 1      & 1      & 2      & 6      & 2      & 2      \\
4     & 1     & 4     & 2     & 1     & 2     & 4     & 5     & 5     & 4      &        &        &        &        &        &        &        &        &        &        \\
5     & 5     & 2     & 4     & 3     & 2     & 4     & 2     & 6     & 1      & 1      & 5      & 2      & 4      & 1      & 3      & 6      & 2      & 1      & 6      \\
2     & 2     & 2     & 2     & 2     & 2     & 2     & 2     & 2     & 2      & 2      & 2      & 2      & 2      & 2      & 2      & 2      & 2      & 2      & 2      \\
2     & 1     & 2     & 2     & 1     & 2     & 2     & 2     & 2     & 2      & 2      & 2      & 2      & 2      & 2      & 1      & 6      & 2      & 2      & 2      \\
2     & 2     & 4     & 2     & 1     & 2     & 2     & 5     & 1     & 2      & 2      & 2      & 2      & 5      & 1      & 2      & 2      & 2      & 2      & 2      \\
2     & 1     & 2     & 2     & 2     & 2     & 2     & 2     & 1     & 2      & 2      & 2      & 2      & 2      & 2      & 1      & 2      & 2      & 4      & 2      \\
2     & 1     & 2     & 2     & 3     & 6     & 2     & 2     & 5     & 6      & 2      & 2      & 2      & 4      & 6      & 3      & 2      & 2      & 2      & 2      \\
4     & 2     & 2     & 2     & 2     & 1     & 2     & 2     & 1     & 2      & 2      & 2      & 2      & 4      & 2      & 2      & 2      & 2      & 2      & 2      \\
2     & 2     & 4     & 2     & 2     & 2     & 2     & 2     & 1     & 1      & 2      & 2      & 2      & 4      & 4      & 4      & 4      & 2      & 2      & 2 \\
\bottomrule
\end{tabular}
\end{adjustbox}
\end{table}

%%%%%%%%%%%%%%%%%%%%%%%%%%%%%%%%%%%%%%%%%%%%%%%%%%%%%%%%%%%%%%
\begin{table}[p] 
\caption{User study results on image quality. Each row shows the choice of one participant. \label{tab:user2}}
\begin{adjustbox}{width=1.0\columnwidth,center}
\begin{tabular}{cccccccccccccccccccc}
\toprule
case1 & case2 & case3 & case4 & case5 & case6 & case7 & case8 & case9 & case10 & case11 & case12 & case13 & case14 & case15 & case16 & case17 & case18 & case19 & case20 \\
\midrule
2     & 2     & 2     & 2     & 2     & 2     & 2     & 2     & 2     & 2      & 2      & 2      & 2      & 2      & 2      & 2      & 2      & 2      & 2      & 2      \\
2     & 2     & 4     & 2     & 2     & 2     & 2     & 2     & 2     & 2      & 2      & 1      & 2      & 1      & 2      & 4      & 1      & 2      & 2      & 2      \\
2     & 2     & 2     & 2     & 2     & 2     & 2     & 2     & 1     & 2      & 2      & 2      & 2      & 2      & 2      & 2      & 2      & 2      & 2      & 2      \\
2     & 4     & 4     & 2     & 2     & 2     & 2     & 2     & 5     & 2      & 2      & 2      & 4      & 4      & 2      & 2      & 2      & 4      & 4      & 2      \\
4     & 2     & 4     & 2     & 2     & 4     & 2     & 2     & 1     & 2      & 2      & 2      & 2      & 4      & 2      & 2      & 2      & 2      & 2      & 2      \\
2     & 2     & 2     & 2     & 2     & 2     & 2     & 1     & 1     & 2      & 2      & 2      & 2      & 4      & 2      & 1      & 1      & 2      & 2      & 2      \\
1     & 2     & 2     & 4     & 2     & 2     & 2     & 2     & 1     & 2      & 2      & 2      & 2      & 4      & 1      & 2      & 2      & 2      & 2      & 1      \\
4     & 4     & 4     & 4     & 4     & 4     & 2     & 4     & 4     & 3      & 2      & 2      & 2      & 4      & 4      & 4      & 2      & 4      & 2      & 2      \\
2     & 2     & 2     & 2     & 2     & 2     & 2     & 2     & 2     & 2      & 2      & 2      & 2      & 4      & 1      & 2      & 4      & 4      & 2      & 2      \\
2     & 2     & 2     & 2     & 2     & 2     & 2     & 2     & 2     & 2      & 2      & 2      & 2      & 2      & 2      & 2      & 2      & 2      & 2      & 2      \\
2     & 2     & 2     & 2     & 2     & 2     & 2     & 2     & 1     & 2      & 2      & 2      & 2      & 2      & 2      & 1      & 2      & 2      & 2      & 2      \\
2     & 2     & 2     & 2     & 2     & 2     & 2     & 2     & 2     & 2      & 2      & 2      & 2      & 4      & 2      & 2      & 2      & 2      & 2      & 2      \\
2     & 2     & 2     & 4     & 2     & 1     & 2     & 2     & 2     & 2      & 2      & 2      & 2      & 4      & 4      & 1      & 2      & 6      & 2      & 2      \\
4     & 4     & 4     & 4     & 4     & 2     & 4     & 4     & 4     & 1      &        &        &        &        &        &        &        &        &        &        \\
2     & 2     & 2     & 2     & 2     & 2     & 2     & 2     & 5     & 3      & 2      & 2      & 2      & 2      & 2      & 1      & 2      & 2      & 1      & 2      \\
2     & 2     & 2     & 2     & 2     & 2     & 2     & 2     & 2     & 2      & 2      & 2      & 2      & 2      & 2      & 2      & 2      & 2      & 2      & 2      \\
2     & 2     & 2     & 2     & 2     & 2     & 2     & 2     & 2     & 2      & 2      & 2      & 2      & 2      & 2      & 2      & 6      & 2      & 2      & 2      \\
2     & 2     & 4     & 2     & 2     & 2     & 2     & 2     & 1     & 1      & 2      & 2      & 2      & 2      & 1      & 2      & 1      & 2      & 2      & 2      \\
2     & 2     & 2     & 2     & 2     & 2     & 2     & 2     & 4     & 2      & 2      & 2      & 2      & 2      & 2      & 2      & 2      & 2      & 2      & 2      \\
2     & 2     & 2     & 2     & 2     & 2     & 2     & 2     & 2     & 2      & 2      & 2      & 2      & 2      & 2      & 2      & 2      & 2      & 2      & 2      \\
2     & 2     & 2     & 2     & 2     & 2     & 2     & 2     & 2     & 2      & 2      & 2      & 2      & 2      & 2      & 2      & 2      & 2      & 4      & 2      \\
2     & 2     & 4     & 2     & 2     & 2     & 2     & 2     & 1     & 1      & 2      & 2      & 2      & 4      & 2      & 4      & 4      & 4      & 2      & 2     \\
\bottomrule
\end{tabular}
\end{adjustbox}
\end{table}

\end{document}